\documentclass[runningheads]{llncs}

\usepackage{epsfig}
\usepackage{amsmath}
\usepackage{amssymb}
\usepackage{multirow}
\usepackage{colortbl}

\usepackage{amsmath}
\usepackage{array}
\usepackage{wrapfig,lipsum,booktabs}
\usepackage{mathalfa}
\usepackage{amssymb}
\definecolor{grey}{rgb}{0.91,0.91,0.91}

\usepackage{graphicx}

\usepackage{tikz}
\usepackage{comment}
\usepackage{amsmath,amssymb} 
\usepackage{color}
\usepackage{orcidlink}

\usepackage[accsupp]{axessibility}  

\begin{document}
\newcommand{\yifan}[1]{{\color{blue} Yifan: #1}}
\newcommand{\tianfan}[1]{{\color{green} Tianfan: #1}}
\newcommand{\bart}[1]{{\color{red} Bart: #1}}
\newcommand{\barron}[1]{{\color{cyan} Jon: #1}}
\newcommand{\ben}[1]{{\color{magenta} Ben: #1}}

\def\ourslong{Malleable Convolution}
\def\ours{MalleConv\,}
\def\ourop{slice-and-conv}
\def\OurOp{Slice-and-Conv}

\pagestyle{headings}
\mainmatter
\def\ECCVSubNumber{3257}  

\title{Fast and High Quality Image Denoising via \\Malleable Convolution} 

\titlerunning{Malleable Convolution}
%
\author{Yifan Jiang\thanks{This work was performed while Yifan Jiang worked at Google.}\inst{1} \and
Bartlomiej Wronski\inst{2} \and 
Ben Mildenhall\inst{2} \and
Jonathan T. Barron\inst{2} \and \\
Zhangyang Wang \inst{1}\and
Tianfan Xue \inst{2} 
}
\authorrunning{Y. Jiang et al.}
%
\institute{University of Texas at Austin \and Google Research }
\maketitle


\begin{abstract}
Most image denoising networks apply a single set of static convolutional kernels across the entire input image. This is sub-optimal for natural images, as they often consist of heterogeneous visual patterns.
Dynamic convolution tries to address this issue by using per-pixel convolution kernels, but this greatly increases computational cost. 
In this work, we present \textbf{Malle}able \textbf{Conv}olution (\textbf{MalleConv}), which performs spatial-varying processing with minimal computational overhead. MalleConv uses a smaller set of spatially-varying convolution kernels, a compromise between static and per-pixel convolution kernels. These spatially-varying kernels are produced by an efficient predictor network running on a downsampled input, making them much more efficient to compute than per-pixel kernels produced by a full-resolution image, and also enlarging the network's receptive field compared with static kernels. These kernels are then jointly upsampled and applied to a full-resolution feature map through an efficient on-the-fly slicing operator with minimum memory overhead. To demonstrate the effectiveness of MalleConv, we use it to build an efficient denoising network we call \textbf{MalleNet}. MalleNet achieves high-quality results without very deep architectures, making it 8.9$\times$ faster than the best performing denoising algorithms while achieving similar visual quality. We also show that a single MalleConv layer added to a standard convolution-based backbone can significantly reduce the computational cost or boost image quality at a similar cost. More information is on our project page: \url{https://yifanjiang.net/MalleConv.html}
\end{abstract}

\section{Introduction}
\begin{figure*}
\centering  
\includegraphics[width=0.7\linewidth]{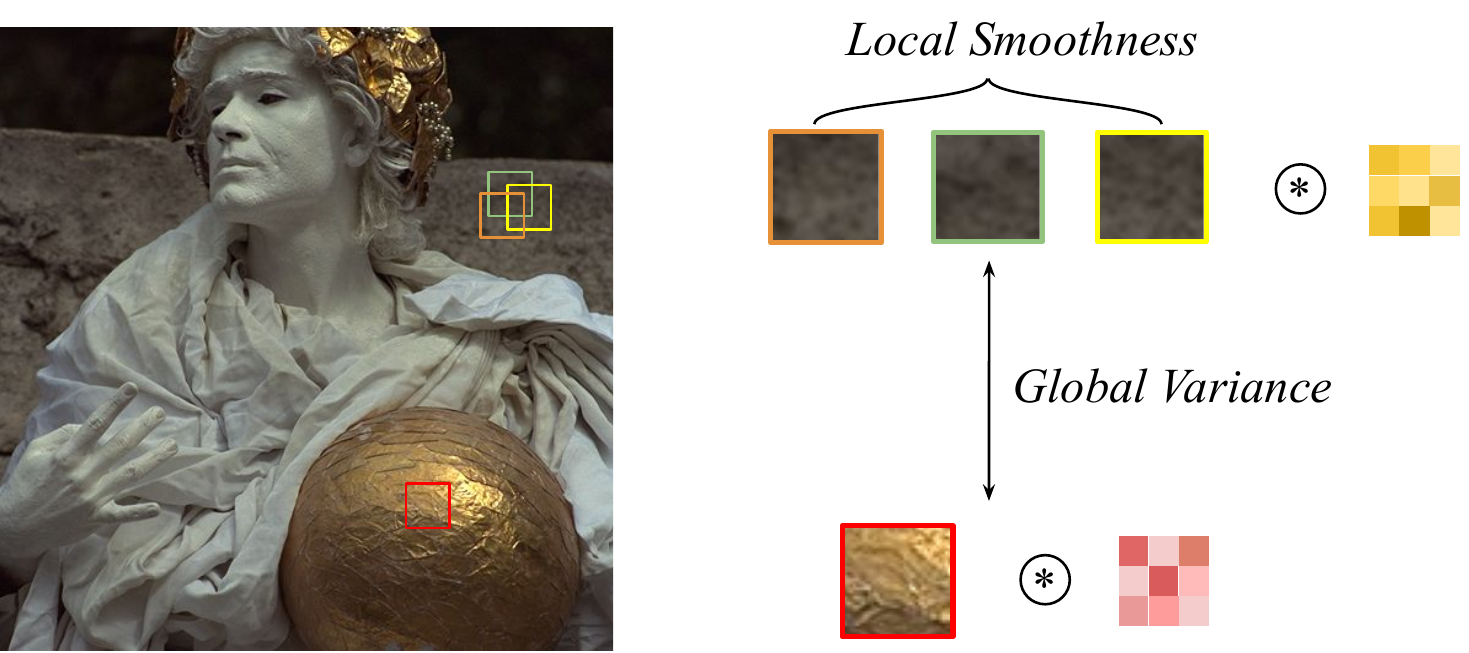}
\caption{\textbf{Local smoothness and global variance in natural images.} Our proposed MalleConv layer applies spatially-varying filters for features in different contexts and adopt similar filters in areas that are locally smooth, thus balancing the trade-off between global variance and local smoothness.}
\label{fig:teaser}
\end{figure*}
Image denoising is a fundamental problem to computational photography and computer vision. Recent advances in deep learning have sparked significant interest in learning an end-to-end mapping directly from corrupted observations to the unobserved clean signal, without an explicit model of signal corruptions. These networks appear to learn a prior over the appearance of ``ground truth'' noise-free images in addition to the statistical properties of the noise present in the inputs.

The performance of denoising networks has consistently been improved with deeper and wider layers, as they can extract richer representations and also increase the receptive field. However, deeper and wider layers also significantly amplify computational costs and the difficulty of optimization. One hurdle is that most of neural architectures only apply a single fixed set of convolutional kernels over the entire input, exploiting spatial equivariance for computational efficiency. However, natural images often contain spatially heterogeneous visual patterns, depriving the convolution of the ability to adapt to globally varying features.

One recent effort addresses this issue is a kernel prediction network (or ``hypernetwork'')~\cite{bako2017kernel,ha2016hypernetworks,jia2016dynamic,mildenhall2018burst,wang2019carafe,xu2020unified}, which generates spatially-varying kernels at each pixel location. 
Although applying per-pixel kernels increases representational power, it also greatly increases computational cost, as the number of kernels grows with the image resolution.
This makes it particularly challenging for mobile cellphone cameras, which normally have about 12 megapixels, and very limited compute resources and power budget.
\begin{figure*}
\centering  
\includegraphics[width=0.99\linewidth]{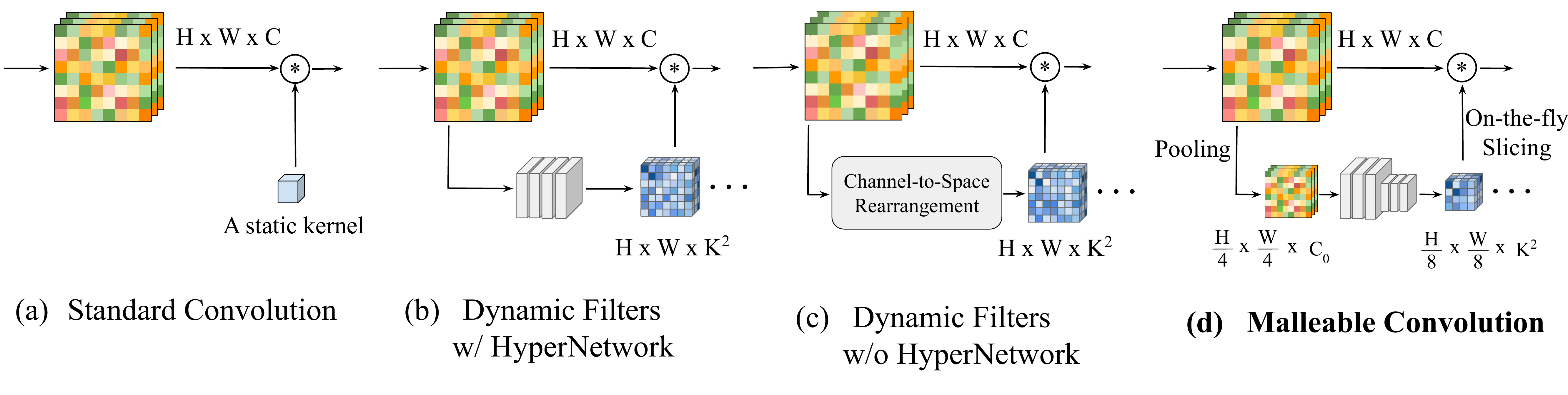}
\caption{\textbf{Comparing MalleConv with static filter and other dynamic filters.} (a) Standard convolution with a static kernel. (b) Generate dynamic filters using a HyperNetwork~\cite{ha2016hypernetworks,jia2016dynamic}. (c) Generate dynamic filters using a channel-to-space operation~\cite{li2021involution}. (d) Our Malleable convolution.}
\label{fig:Difference}
\end{figure*}

To achieve spatial-varying processing while maintaining low computational cost, we propose an efficient variant of spatially-varying kernels, dubbed Malleable Convolution (\textbf{MalleConv}). We draw inspiration from the trade-off between local smoothness and global spatial heterogeneity. Fundamentally, natural images contain spatially-varying patterns from a ``global'' perspective, which motivates the popularity of dynamic filters~\cite{ha2016hypernetworks,jia2016dynamic} and self-attention modules~\cite{liang2021swinir,vaswani2017attention}, but image content only changes slowly in a ``local'' neighborhood. Therefore, natural image patches tend to redundantly recur many times inside the image, both within the same scale and across different scales~\cite{peyre2008non,glasner2009super}. Natural image textures are also commonly represented as a fractal set with self-similarity at all scales \cite{kataoka2020pre}. Examples in Fig.~\ref{fig:teaser} also illustrate this phenomenon. The golden ball held by the man contains different patterns compared to the stone in the background, but the texture is locally consistent within a region of stone. Therefore, those similar content can share the same set of kernels to save compute.

Based on this observation, we proposed MalleConv, which scales per-pixel dynamic filter approach to a larger region. Specifically, unlike dynamic filters which take full-resolution input and generate full-resolution kernels, MalleConv only processes a downsampled representation, outputting location-specific dynamic filters at \textbf{a much smaller spatial resolution} compared with the original feature map (Fig.~\ref{fig:Difference}(d)). These kernels are later applied to the full-resolution feature map using a ``slicing'' strategy, which fuses on-the-fly bilinear interpolation and convolution into a single operator. This design has several advantages. First, comparing to the hypernetwork used in dynamic filters, our predictor network only takes a low-resolution feature map as input to keep it light-weight. Second, full resolution per-pixel kernels are calculated and applied in the same operation, without requiring additional memory I/O for storing and retrieving the high resolution kernel map. Together, these significantly reduce computational overhead compared to full-resolution dynamic filters. Moreover, by taking a downsampled image as input, the predictor network has a large receptive field without very deep structure. 

Comprehensive experiments are conducted to demonstrate the effectiveness of the proposed method. We evaluate MalleNet on public synthetic and real image benchmarks (Synthetic: CBSD68, Kodak24, McMaster; Real: SIDD and DND). In addition, we conduct ablation study by injecting MalleConv into existing backbones, including DnCNN~\cite{zhang2017beyond}, UNet, and RDN~\cite{zhang2018residual}, where the results show that MalleConv achives better quality-efficiency trade-off compared to other dynamic kernels.


In summary, our contributions are as follows: 
\begin{itemize}
\setlength{\itemsep}{0pt}
\setlength{\parskip}{0pt}
    \item We propose Malleable Convolution (MalleConv), a new spatially-varying kernel layer that serves as a powerful variant of standard convolution. MalleConv largely benefits from an efficient predictor network, which incurs minimum additional cost to achieve a spatial-varying processing.
    \item We conduct a comprehensive ablation study by inserting MalleConv into various popular backbone architectures (including DnCNN, UNet, and RDN), where we show MalleConv can reduce runtime by up to \textbf{20}$\times$ with similar visual quality.
    \item We compare MalleConv with previous spatially-varying kernel architectures including HyperNetworks~\cite{ha2016hypernetworks} and Involution~\cite{li2021involution}. MalleConv demonstrates a better quality-efficiency trade-off.
    \item We further design a new MalleNet architecture using the proposed MalleConv block, achieving faster performance and higher quality on both synthetic and real-world denoising benchmarks.
\end{itemize}

\section{Related Work}
\subsection{Image Denoising}
Traditional image denoising algorithms make use of information in local pixel neighborhoods~\cite{perona1990scale,rudin1992nonlinear} or sparse image prior~\cite{aharon2006k,elad2006image,mairal2009non,buades2005non,getreuer2018blade,dabov2007image}.
Recently, deep convolutional networks have demonstrated success in many image restoration tasks~\cite{dong2015image,lim2017enhanced,zhang2018residual,tao2018scale,kupyn2018deblurgan,kupyn2019deblurgan,ren2019progressive,li2019single,yang2020single,jiang2021enlightengan,wei2018deep,chen2022cerl,jiang2021ssh}. For image denoising specifically, Burger et al.~\cite{burger2012image} proposed a plain multi-layer perception model that achieves comparable performance to BM3D. Chen et al.~\cite{chen2016trainable} proposed a trainable nonlinear reaction diffusion model that learns to remove additive white gaussian noise (AWGN) by unfolding a fixed number of inference steps. Many subsequent works further improved upon it by using more elaborate neural network architecture designs, including residual learning~\cite{zhang2017beyond}, dense networks~\cite{zhang2018residual}, non-local modules~\cite{zhang2019residual,chen2021pre,liang2021swinir}, dilated convolutions~\cite{peng2019dilated}, and more~\cite{cheng2021nbnet,zamir2021multi,zamir2020learning,chen2021hinet}. However, many of these approaches use heavy network architectures that are often impractical for mobile use cases. To tackle this issue, several recent works focus on fast image denoising, by either introducing a self-guidance network~\cite{gu2019self} or increasing the nonlinear model capacity~\cite{gu2019fast}. In contrast, our approach relies on spatially-varying kernels, where parameters are dynamically generated by an efficient prediction network. 

\subsection{Dynamic Filters and Spatially Varying Kernels}
Convolutional neural networks producing dynamic kernels have been widely studied for a variety of applications. The pioneering works~\cite{jia2016dynamic,ha2016hypernetworks} adopt a parameter-generating network to produce location-specific filters.  These works directly produce spatially-varying weights for the whole convolutional layer, substantially increasing the latency and computational cost of their approaches. Wang et al.~\cite{wang2019carafe} designed a feature upsampling module (CARAFE) that generates kernels and reassembling features inside a predefined nearby region. However, CARAFE is designed as a feature upsampling operator instead of a variant of convolution. The context-gated convolution~\cite{lin2020context,zhang2021context} adopts a gated module and channel/spatial interaction module to generate modified convolutional kernels. Although their filter weights are produced dynamically, they apply the same filter at different spatial locations. Another line of work~\cite{li2021involution} avoids using a hypernetwork by employing a channel-to-space rearrangement to generate location-specific filters. Without the help of a hypernetwork, this approach can not capture the local information and image context. While previously described approaches mainly adopt dynamic filters inside multiple convolutional layers of a deep network, a different line of work\cite{bako2017kernel} proposed to use a standard convolutional neural network to predict denoising kernels that are applied directly to the input to produce the target image. 
Mildenhall et al.~\cite{mildenhall2018burst} extended this approach to burst denoising by predicting a separate set of weights for each image in a temporal sequence. 
HDRNet~\cite{gharbi2017deep} uses a deep neural network to process the low-resolution input and applies the produced spatially-varying affine matrix to the full-resolution input by slicing a predicted bilateral grid. In stead of processing the input image, our proposed Malleable Convolution applies an efficient predictor network to process a downsampled feature map, then constructs a deep spatially-varying network layer-by-layer. 

\begin{figure*}[t]
\centering
\resizebox{0.99\linewidth}{!}{
\includegraphics[]{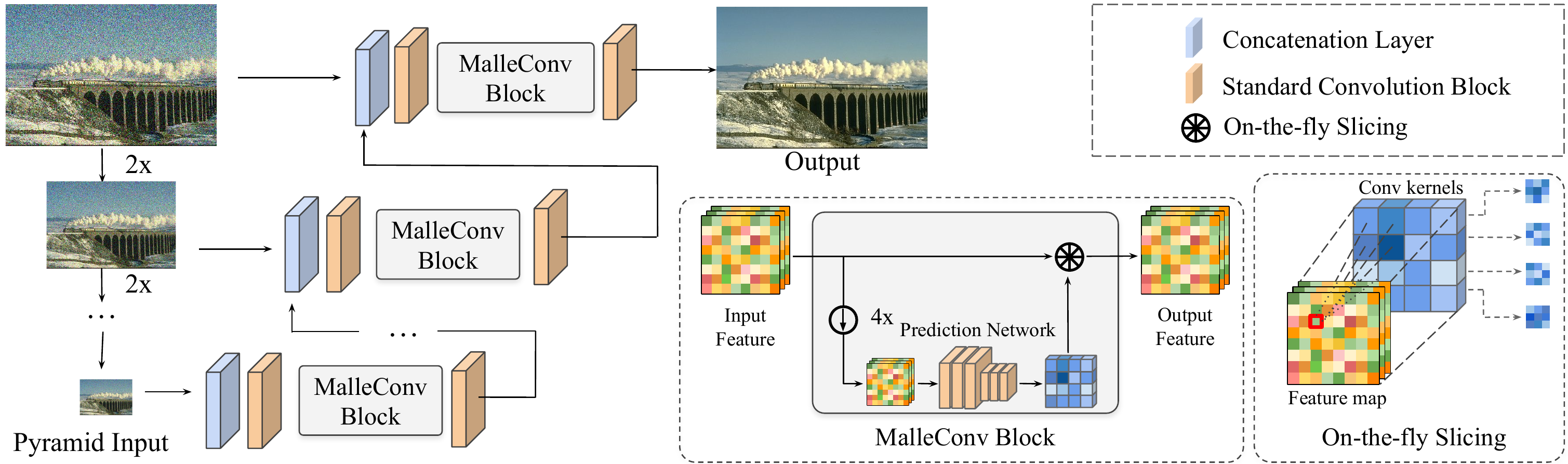}
}
\caption{\textbf{Main architecture of MalleNet.} MalleNet takes a 4-level image pyramid as input. 
Each layer consists of several Inverted Bottleneck Blocks with a MalleConv block inserted in between. 
Bottom middle shows the structure of MalleConv block, which consists of a small prediction network and a on-the-fly slicing operator. Bottom right shows details of on-the-fly slicing operator.
For each input feature (red rectangle), four neighboring kernels are bilinearly combined and applied to that feature to generate the corresponding output feature.}
\label{fig:main_pipeline}
\end{figure*}

\section{Method}
\subsection{Preliminaries}
A standard convolutional layer applies a kernel with weights $ W \in \mathbb{R}^{C_{in} \times C_{out} \times K^2}$ to an input feature map sampled from a 2D tensor $X \in \mathbb{R}^{C_{in} \times H \times W}$. Here $H, W$ are the height and width of the feature map, $C_{in}, C_{out}$ denote the numbers of input and output channels, and $K$ is the kernel size. 
This basic design struggles to capture global context information and cannot adapt to different regions of natural images that contain spatially heterogeneous patterns. Although previous works address this issue by adopting per-pixel dynamic filters~\cite{ha2016hypernetworks,jia2016dynamic,mildenhall2018burst} or generating spatial-agnostic filters via a channel-to-space permutation~\cite{li2021involution}, their approaches either require large memory footprint or do not capture context information.


\subsection{Malleable Convolution with Efficient Predictor Network}
To overcome the aforementioned drawbacks, we propose a new operation, dubbed Malleable Convolution (MalleConv). MalleConv is equipped with an light-weight predictor network that significantly reduces the memory cost and runtime latency of previous dynamic kernel prediction~\cite{ha2016hypernetworks,jia2016dynamic,mildenhall2018burst}. The proposed predictor network 
first downsamples the input feature map $X$ to $X^{'} \in \mathbb{R}^{\frac{H}{4} \times \frac{W}{4} \times C}$ through a $4 \times4$ average pooling. 
After that, we build a light-weight predictor network consists of multiple ResNet blocks~\cite{he2016deep} and max pooling layers~\cite{he2015delving} (see supplementary materials for detailed architecture). The predictor network outputs a feature map $Y \in \mathbb{R}^{\frac{H}{8} \times \frac{W}{8} \times C^{'}}$, where $C^{'}=K^{2} \times C$. To formulate a spatially-varying filter, the learned representation $Y$ is reshaped to a list of filters $ \{W_{ij}\} \in \mathbb{R}^{K^{2} \times C}$, where $i \in \{1, 2, ...,  \frac{H}{8}\}$, $j \in \{1, 2, ..., \frac{W}{8}\}$. Each kernel in $Y$ only has $C$ channels, not $C_{in} \times C_{out}$, as we use depth-wise convolution~\cite{howard2017mobilenets} to further reduce the number of parameters. Finally, we upsample the learned spatially-varying filters $ \{W_{ij}\}$ through bilinear interpolation to obtain per-pixel filters $ \{W^{'}_{ij}\} \in \mathbb{R}^{K^{2} \times C}$, where $i \in \{1, 2, ...,  H\}$, $j \in \{1, 2, ..., W\}$, and independently apply them to the corresponding input channels.

\subsection{Efficient On-the-fly Slicing}
A naive way to implement malleable convolution is to first upsample the low-resolution filters to full-resolution using bilinear interpolation and then apply them to the full-resolution feature map. However, this introduces a large memory footprint since the high-resolution kernels are being precomputed and stored before their application.

To mitigate the memory issue, we combine these two steps into a on-the-fly slicing operator. It takes in a high-resolution feature map $X \in \mathbb{R}^{H \times W \times C}$ and low-resolution kernel maps $ \{W_{ij}\} \in \mathbb{R}^{K^{2} \times C}$ as input. The result of the on-the-fly slicing operator is a new feature map $Z$ with the same resolution as $X$. 
For each pixel location, we first calculate the bilinear interpolated kernel weights from four neighboring kernels as (also illustrated in bottom right of Fig.~\ref{fig:main_pipeline})
\begin{align}
    W^{'}_{x,y} = \sum_{i,j\in N(x,y)} \tau(r_xx - i)\tau(r_yy - j)W{'}_{i,j},
\end{align}
where $\tau$ is the linear interpolation operator $\tau(a) = max(1 - \lvert a \rvert, 0)$, $r_x$ and $r_y$ are the width and height ratios of the low-resolution filters w.r.t. the full resolution input feature map, and $N(x,y)$ is the four-neighborhood. Bias term $b^{'}_{x,y}$ is sliced in the similar way. The output feature $Z$ is then calculated as:
\begin{align}
Z_{x,y}(c) = W^{'}_{x,y}(c) \cdot X_{x,y}(c) + b^{'}_{x,y}(c),
\end{align}
where $c$ is the channel index. Note that the sliced weight $W^{'}$ and bias $b^{'}$ are calculate on-the-fly without additional memory cost.
We discuss more about the specific memory consumption in Sec. \ref{sec:memory_analysis}.



\begin{table*}[t]
\small
\begin{center}
\small
\resizebox{\linewidth}{!}{
\setlength{\tabcolsep}{0.4mm}{
\begin{tabular}{l|c|c|c|c|c|c|c|c|c|c|c} 
\toprule
\multirow{2}{*}{Method}   & \multirow{2}{*}{Latency/(ms)} &\multirow{2}{*}{Flops/(G)} & \multicolumn{3}{c|}{CBSD68} & \multicolumn{3}{c|}{Kodak24} & \multicolumn{3}{c}{McMaster} \\
\cline{4-12}
& &  & $\sigma=15$ & $\sigma=25$ & $\sigma=50$ & $\sigma=15$ & $\sigma=25$ & $\sigma=50$ & $\sigma=15$ & $\sigma=25$ & $\sigma=50$ \\
\hline
BM3D~\cite{dabov2007image}       &   41.56    &   -    & 33.52 & 30.71 & 27.38 & 34.28 & 32.15 & 28.46 & 34.06 & 31.66 & 28.51 \\

FFDNet~\cite{zhang2018ffdnet}    & - & 7.95 & 33.87 & 31.21 & 27.96 & 34.63 & 32.13 & 28.98 & 34.66 & 32.35 & 29.18 \\
\rowcolor{grey}  \textbf{MalleNet-S}              &  \textbf{4.62} & \textbf{2.93} & \textbf{33.90} & \textbf{33.22} & \textbf{27.97} & \textbf{34.66} & \textbf{32.16} & \textbf{29.00} & \textbf{34.68} & \textbf{32.35}  & \textbf{29.20} \\
\hline
RPCNN~\cite{xia2020identifying}  & 95.11 & - &   -   & 31.24 & 28.06 &   -   & 32.34 & 29.25 &   -   & 32.33 & 29.33 \\
DSNet~\cite{peng2019dilated}     & - & - & 33.91 & 31.28 & 28.05 & 34.63 & 32.16 & 29.05 & 34.67 & 32.40 & 29.28 \\
IRCNN~\cite{zhang2017learning}   &    -   &  12.18  & 33.86 & 31.16 & 27.86 & 34.69 & 32.18 & 28.93 & 34.58 & 32.18 & 28.91 \\
DnCNN~\cite{zhang2017beyond}     & 21.69 & 68.15 & 33.90 & 31.24 & 27.95 & 34.60 & 32.14 & 28.95 & 33.45 & 31.52 & 28.62 \\
DnCNN*~\cite{zhang2017beyond}     & 21.69 & 68.15 & 34.02 & 31.34 & 28.11 & 34.62 & 32.18 & 29.11 & 35.18 & 32.73 & 29.49 \\
\rowcolor{grey} \textbf{MalleNet-M}              &  \textbf{16.69} & \textbf{9.36} & \textbf{34.15} & \textbf{31.50} & \textbf{28.27} & \textbf{34.82} & \textbf{32.41} & \textbf{29.35} & \textbf{35.53} & \textbf{33.12}  & \textbf{29.96} \\
\hline

BRDNet~\cite{tian2020image}      & - & - & 34.10 & 31.43 & 28.16 & 34.88 & 32.41 & 29.22 & 35.08 &    32.75  & 29.52 \\

DRUNet~\cite{zhang2021plug}      & - & 102.91 & 34.30 & 31.69 & 28.51 & \textbf{35.31} & \textbf{32.89} & \textbf{29.86} & 35.40 & 33.14  & 30.08 \\
\rowcolor{grey} \textbf{MalleNet-L}            & \textbf{32.34} & \textbf{33.47} & \textbf{34.32} & \textbf{31.71} & \textbf{28.52} & 34.93 & 32.58 & 29.50 & \textbf{35.65} & \textbf{33.26}  & \textbf{30.12} \\
\hline

RNAN~\cite{zhang2019residual}    & - &  774.67 &   -   & -     & 28.27 & -     &   -   & 29.58 & -     &    -     & 29.72 \\
RDN~\cite{zhang2018residual}     & 263.03 & 2001.86 & -     & -     & 28.31 & -     & -     & 29.66 & -     &  -       & - \\
RDN*~\cite{zhang2018residual}     & 263.03 & 2001.86 & 34.29     & 31.69     & 28.37 & 34.89     & 32.52     & 29.68 & 35.55     &  33.16       & 29.92 \\
IPT~\cite{chen2021pre}           & - & 938.66 & -     & -     & 28.39 & -     & -     & 29.64 & -     & -        & 29.98 \\
SwinIR~\cite{liang2021swinir}    & 780.61 & 788.10 &  34.42 & 31.78 & 28.56 & \textbf{35.34} & \textbf{32.89} & \textbf{29.79} & 35.61 & 33/20  & 30.22 \\
\rowcolor{grey} \textbf{MalleNet-XL}            & \textbf{87.55} & \textbf{181.89} & \textbf{34.54} & \textbf{31.86} & \textbf{28.62} & 35.07 & 32.67 & 29.61 & \textbf{35.72} & \textbf{33.28}  & \textbf{30.23} \\
\bottomrule

\end{tabular}}}
\end{center}
\caption{\textbf{Comparing MalleNet with the state-of-the-art methods on three common benchmarks.}  We try our best to use the official implementation provided by the authors to calculate FLOPs and latency. ``*'' denotes that the original methods were trained with small-scale dataset and we retrain these networks with more training data and larger patch size, for fair comparison.}
\label{table:comparison}
\end{table*}

\begin{figure*}[t]
\centering
\resizebox{0.95\linewidth}{!}{
\begin{tabular}[!t]{cccc}
  \includegraphics[trim={0 0 0 0},clip,width=0.3\linewidth]{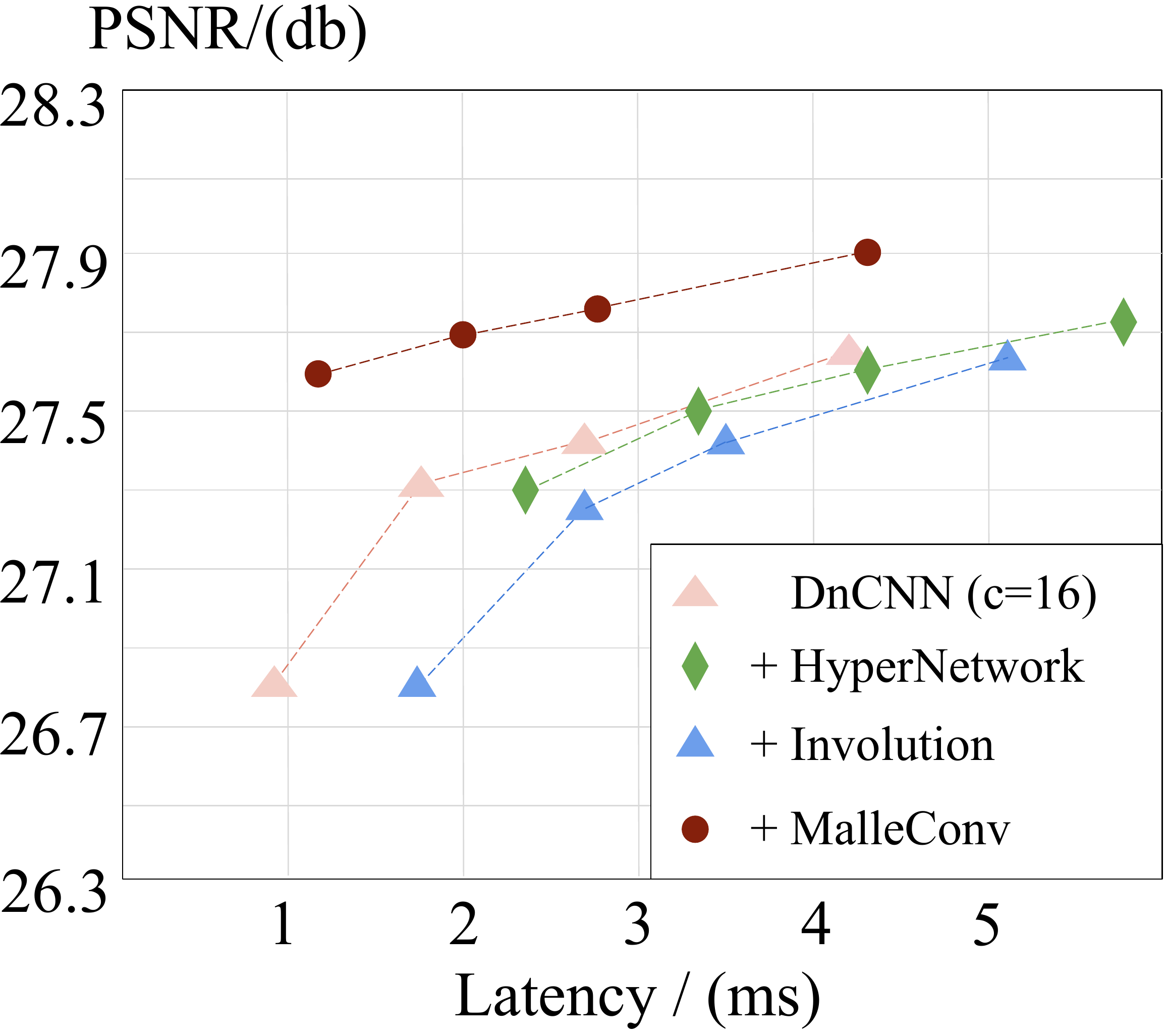} & 
  \includegraphics[trim={0 0 0 0},clip,width=0.3\linewidth]{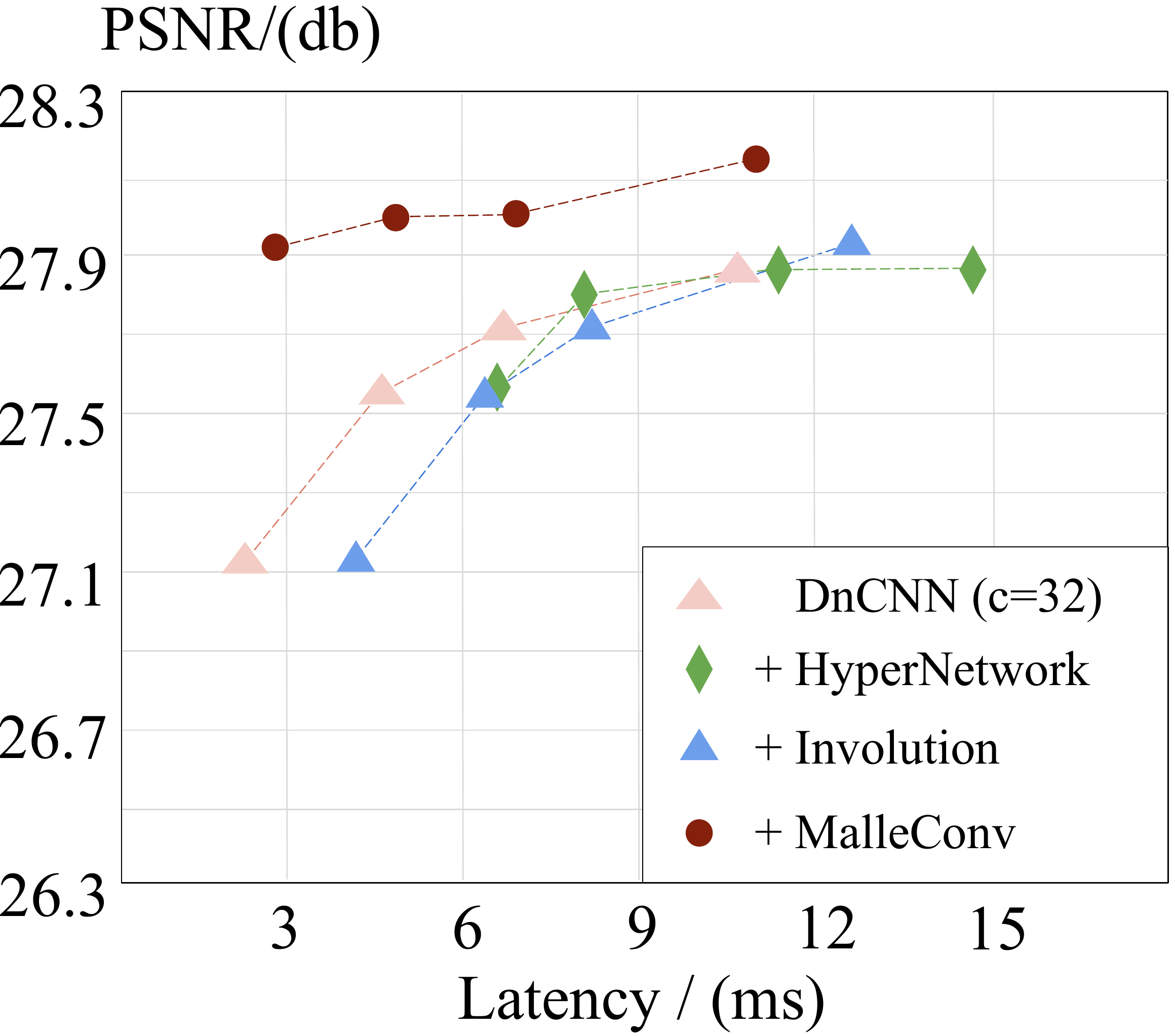} & 
  \includegraphics[trim={0 0 0 0},clip,width=0.3\linewidth]{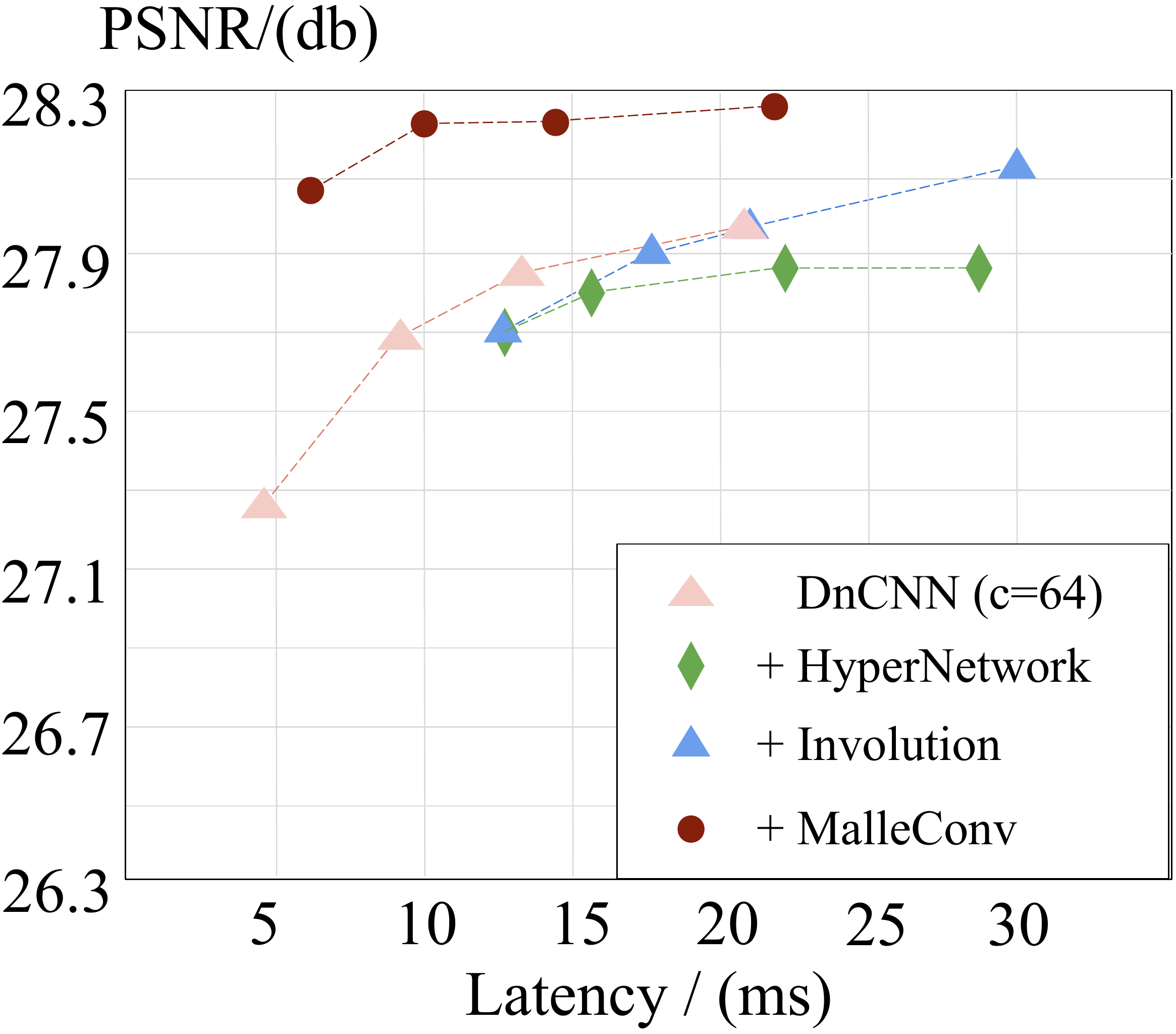} & 
\end{tabular}
}
\caption{Comparison between MalleConv and other dynamic filters in terms of runtime latency and PSNR value.}
\label{fig:hyper_inv_malle}
\end{figure*}

\begin{figure*}[t]
\centering
\resizebox{0.99\linewidth}{!}{
\begin{tabular}[!t]{cccccc}
  \includegraphics[trim={0 0 0 0},clip,width=0.2\linewidth]{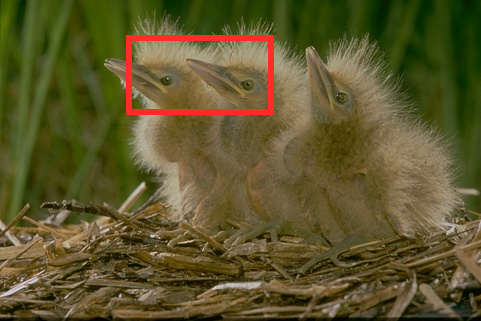} &  \hspace{-1.em}
  \includegraphics[trim={0 0 0 0},clip,width=0.2\linewidth]{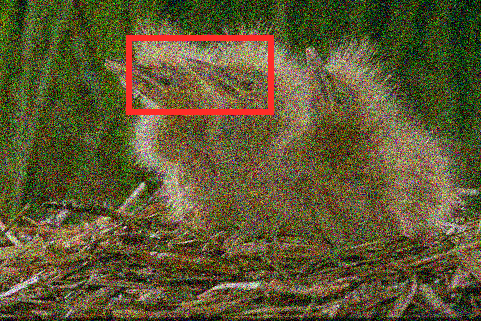} &  \hspace{-1.em}
  \includegraphics[trim={0 0 0 0},clip,width=0.2\linewidth]{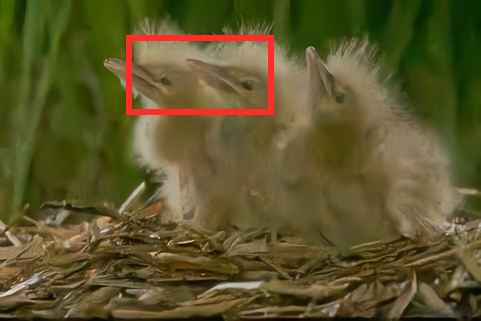} &\hspace{-1.em}
  \includegraphics[trim={0 0 0 0},clip,width=0.2\linewidth]{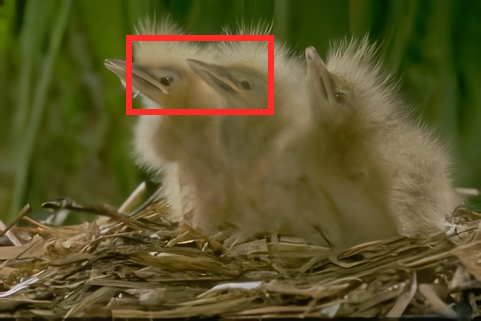} &\hspace{-1.em}
  \includegraphics[trim={0 0 0 0},clip,width=0.2\linewidth]{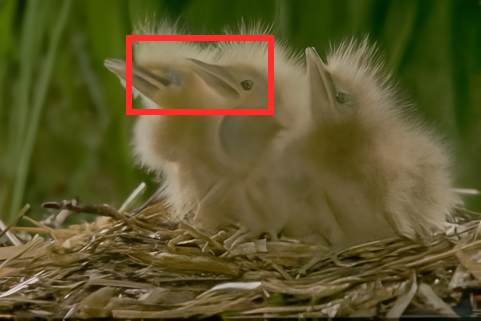} & \hspace{-1.em}
  \includegraphics[trim={0 0 0 0},clip,width=0.2\linewidth]{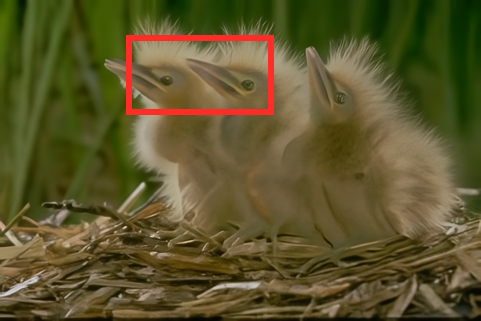} \hspace{-1.em} \\
  \includegraphics[trim={0 0 0 0},clip,width=0.2\linewidth]{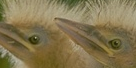} &  \hspace{-1.em}
  \includegraphics[trim={0 0 0 0},clip,width=0.2\linewidth]{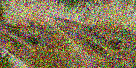} &  \hspace{-1.em}
  \includegraphics[trim={0 0 0 0},clip,width=0.2\linewidth]{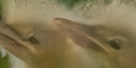} &\hspace{-1.em}
  \includegraphics[trim={0 0 0 0},clip,width=0.2\linewidth]{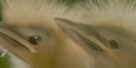} &\hspace{-1.em}
  \includegraphics[trim={0 0 0 0},clip,width=0.2\linewidth]{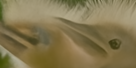} & \hspace{-1.em}
  \includegraphics[trim={0 0 0 0},clip,width=0.2\linewidth]{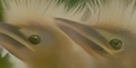} \hspace{-1.em} \\
  
  \includegraphics[trim={0 0 0 0},clip,width=0.2\linewidth]{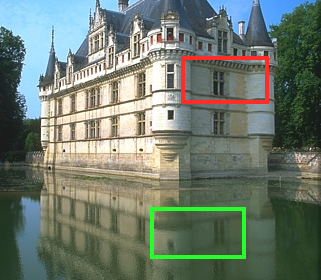} &  \hspace{-1.em}
  \includegraphics[trim={0 0 0 0},clip,width=0.2\linewidth]{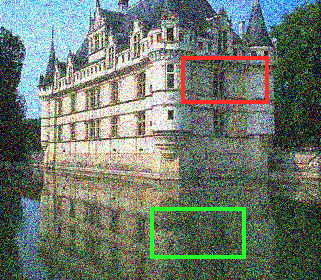} &  \hspace{-1.em}
  \includegraphics[trim={0 0 0 0},clip,width=0.2\linewidth]{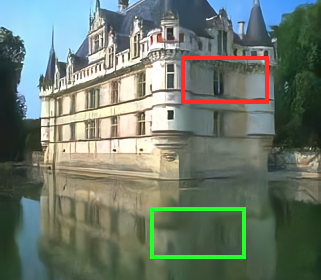} &\hspace{-1.em}
  \includegraphics[trim={0 0 0 0},clip,width=0.2\linewidth]{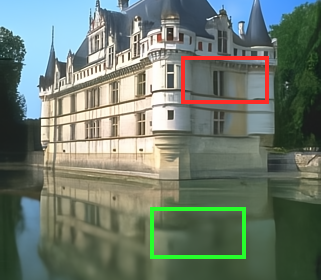} &\hspace{-1.em}
  \includegraphics[trim={0 0 0 0},clip,width=0.2\linewidth]{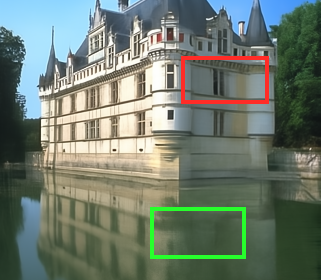} & \hspace{-1.em}
  \includegraphics[trim={0 0 0 0},clip,width=0.2\linewidth]{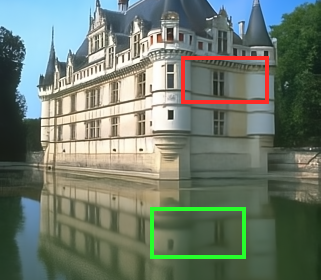} \hspace{-1.em} \\
  \includegraphics[trim={0 0 0 0},clip,width=0.2\linewidth]{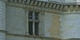} &  \hspace{-1.em}
  \includegraphics[trim={0 0 0 0},clip,width=0.2\linewidth]{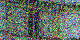} &  \hspace{-1.em}
  \includegraphics[trim={0 0 0 0},clip,width=0.2\linewidth]{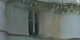} &\hspace{-1.em}
  \includegraphics[trim={0 0 0 0},clip,width=0.2\linewidth]{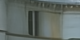} &\hspace{-1.em}
  \includegraphics[trim={0 0 0 0},clip,width=0.2\linewidth]{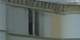} & \hspace{-1.em}
  \includegraphics[trim={0 0 0 0},clip,width=0.2\linewidth]{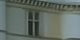} \hspace{-1.em} \\
  \includegraphics[trim={0 0 0 0},clip,width=0.2\linewidth]{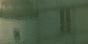} &  \hspace{-1.em}
  \includegraphics[trim={0 0 0 0},clip,width=0.2\linewidth]{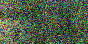} &  \hspace{-1.em}
  \includegraphics[trim={0 0 0 0},clip,width=0.2\linewidth]{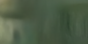} &\hspace{-1.em}
  \includegraphics[trim={0 0 0 0},clip,width=0.2\linewidth]{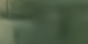} &\hspace{-1.em}
  \includegraphics[trim={0 0 0 0},clip,width=0.2\linewidth]{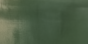} & \hspace{-1.em}
  \includegraphics[trim={0 0 0 0},clip,width=0.2\linewidth]{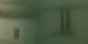} \hspace{-1.em} \\
  GT & Noisy & DnCNN~\cite{zhang2017beyond} & RDN~\cite{zhang2018residual} & SwinIR~\cite{liang2021swinir} & MalleNet
  
\end{tabular}
}
\caption{Visual comparison between MalleNet and previous approaches. More visual results are shown in the supplementary.}
\label{fig:visual_comparison}
\end{figure*}

\subsection{Malleable Network}
As the goal of this work is to design an ultra-fast denoiser, current state-of-the-art algorithms such as the residual dense network~\cite{zhang2018residual} or transformer-based architectures~\cite{chen2021pre,liang2021swinir} are sub-optimal to build an efficient backbone. Inspired by some recent pyramid-based approaches~\cite{gu2019self,liang2021high,zamir2020learning}, we design a new backbone integrating the proposed malleable convolution, dubbed \textbf{MalleNet}. 

MalleNet first builds a four-level pyramid using $2 \times$ space-to-channel shuffle operations~\cite{shi2016real}. This allows us to extract multi-scale representations and increases the network's receptive field. In each stage, we stack several Inverted Bottleneck Blocks~\cite{sandler2018mobilenetv2} with a fixed ratio and insert one $K \times K$ Malleable Convolution in-between to extract heterogeneous representations. At the end of the bottom stage, we upsample the feature map and concatenate it with the input of its upper stage. In the top stage, the representation extracted from different pyramids are aggregated to produce the final output. Compared to conventional encoder-decoder style architectures, the pyramid-based architecture reuses the extracted representation from each scale and thus can achieve faster inference speed. The whole network is shown in Fig.~\ref{fig:main_pipeline}.

\section{Experiments}
We mainly evaluate the proposed module on the Additive White Gaussian Noise (AWGN) removal task. Following previous work~\cite{zhang2021plug}, we construct a training dataset with 400 examples from the Berkeley Segmentation Dataset (BSD)~\cite{MartinFTM01}, 4,744 examples from the Waterloo Exploration Database~\cite{ma2016waterloo}, 900 images from the DIV2K dataset~\cite{agustsson2017ntire}, and 2,750 images from the Flick2K dataset~\cite{lim2017enhanced}. We adopt $160 \times 160$ training patch size, which we augment through random cropping, rotations, and flipping. Other networks (e.g., IPT~\cite{chen2021pre}  and SwinIR~\cite{liang2021swinir}) are not able to be benefited from larger patch size, due to the heavy memory cost. We empirically choose kernels size $1 \times 1$ for MalleConv on AWGN removal tasks and kernel size $3 \times 3$ on real-world benchmarks, as that is observed to reach the best PSNR-to-Complexity trade-off. We adopt the Adam optimizer~\cite{adam} with a batch size of 16 and a cosine learning rate scheduler. The initial learning rate is set to $0.001$. The full training process takes 2.2M iterations. We adopt 3 common datasets as our testing set: CBSD68, kodak24~\cite{franzen1999kodak}, and McMaster~\cite{zhang2011color}. 

All of our experiments are conducted on 8 Nvidia V100 GPUs using the Tensorflow-2.6 platform. The FLOPs (floating point operations) and runtime are calculated on a $256 \times 256$ resolution RGB patches. We benchmar the  inference speed on a single Nvidia P6000 GPU platform by setting batch size set to the maximum available number. For PyTorch-based implementations, we report the average latency of a single $256 \times 256 \times 3$ input collected from 500 runs. For Tensorflow-based implementations, we report the latency time using the Tensorflow official profiler\footnote{https://www.tensorflow.org/guide/profiler}.

\begin{figure}[t]
\centering
\resizebox{0.85\linewidth}{!}{
\begin{tabular}[!t]{@{}c@{\,\,\,}c@{}}
  

  \includegraphics[trim={0 0 0 0},clip,height=0.25\linewidth]{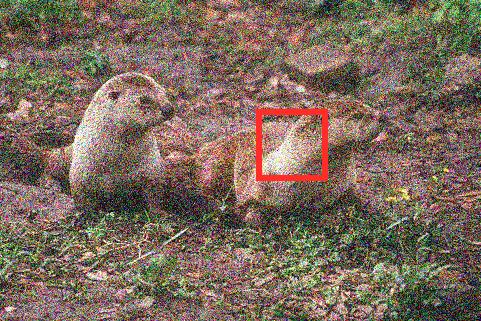}
  \includegraphics[trim={0 0 0 0},clip,height=0.25\linewidth]{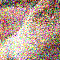}
  &  
  \includegraphics[trim={0 0 0 0},clip,height=0.25\linewidth]{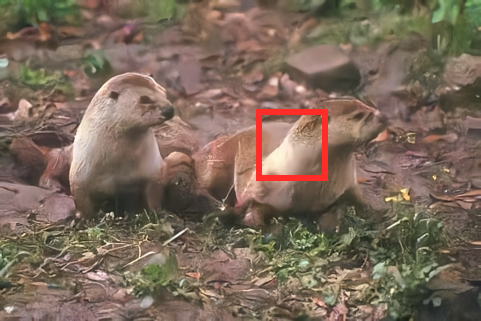}
  \includegraphics[trim={0 0 0 0},clip,height=0.25\linewidth]{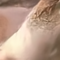}
  \\
    \multirow{2}{*}{Noisy}  & DnCNN (Depth=15) \\
    & latency: \textbf{21.69} ms \\
  \includegraphics[trim={0 0 0 0},clip,height=0.25\linewidth]{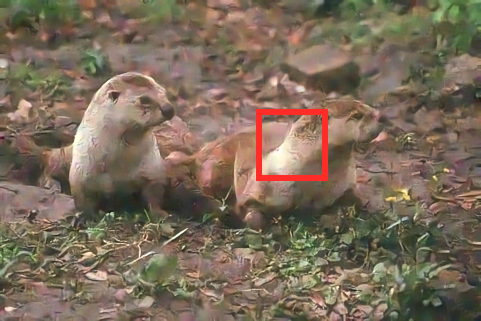} 
    \includegraphics[trim={0 0 0 0},clip,height=0.25\linewidth]{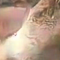} 
  &  
  \includegraphics[trim={0 0 0 0},clip,height=0.25\linewidth]{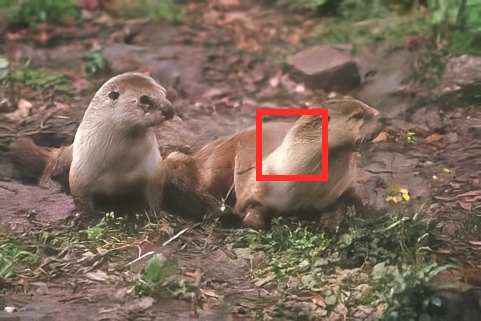}
  \includegraphics[trim={0 0 0 0},clip,height=0.25\linewidth]{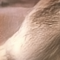}   
  \\
  DnCNN (Depth=3) &  DnCNN (Depth=2) + MalleConv \\
  Latency: \textbf{4.62} ms & Latency: \textbf{5.81} ms 
\end{tabular}
}
\caption{Visual results by inserting MalleConv into a fast variant of DnCNN, with $\sigma$ = 50.}
\label{fig:dncnn_visual_comparison}
\end{figure}

\begin{figure*}
\center
\begin{minipage}[t]{\textwidth}
  \resizebox{0.99\linewidth}{!}{
\begin{tabular}[!t]{ccc}
  \includegraphics[trim={0 0 0 0},clip,width=0.33\linewidth]{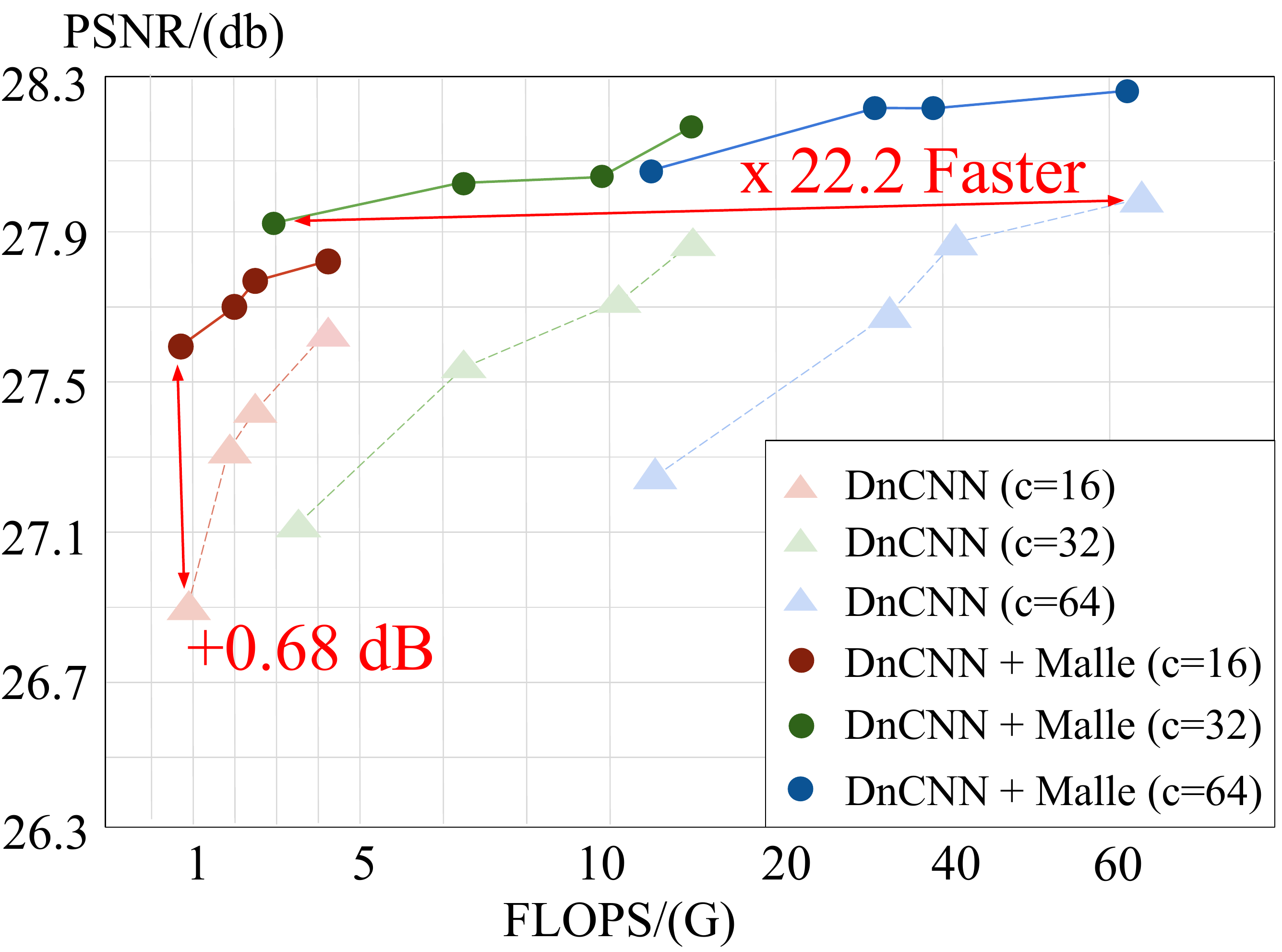} &  
  \includegraphics[trim={0 0 0 0},clip,width=0.33\linewidth]{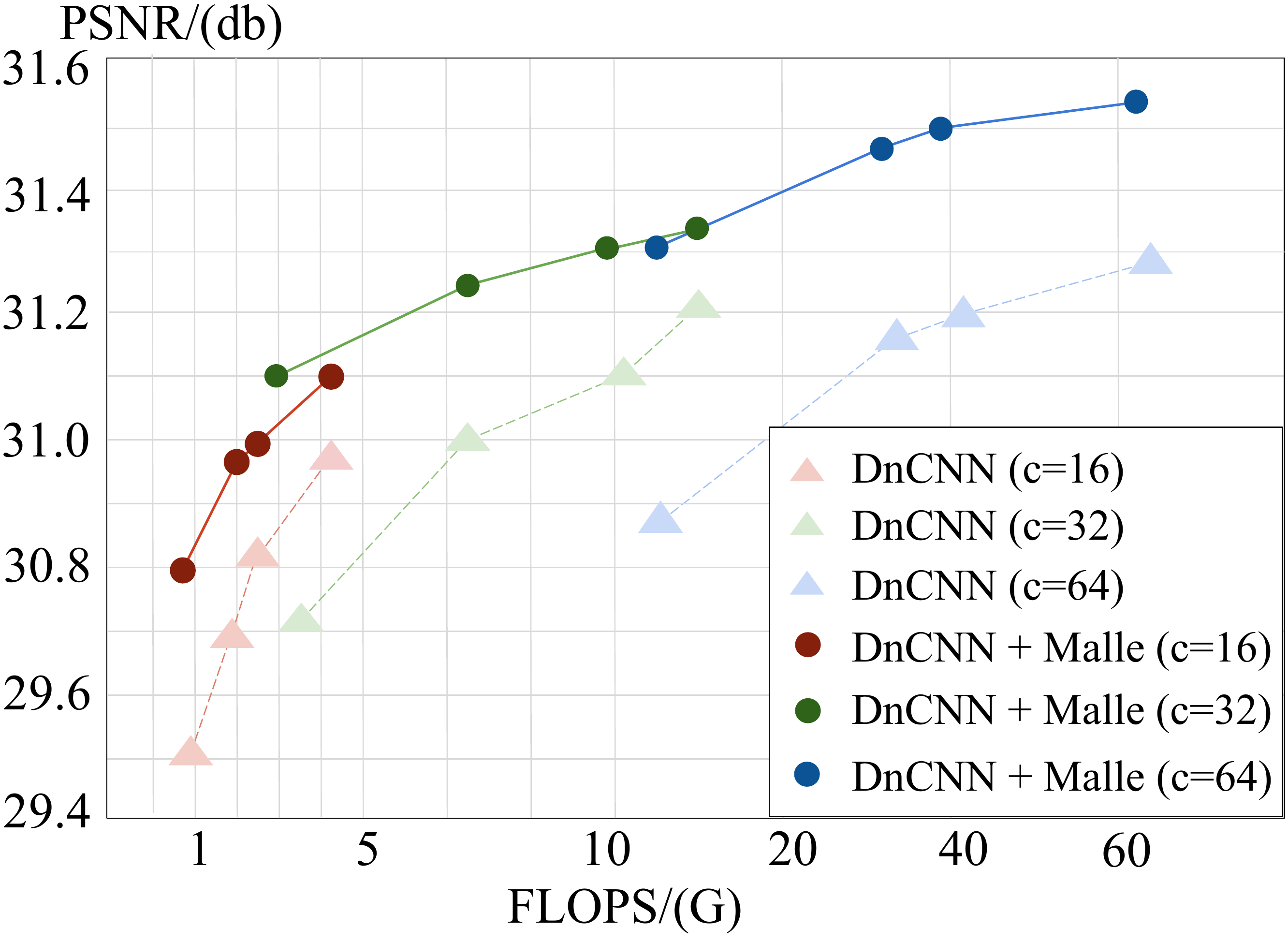} &  
  \includegraphics[trim={0 0 0 0},clip,width=0.33\linewidth]{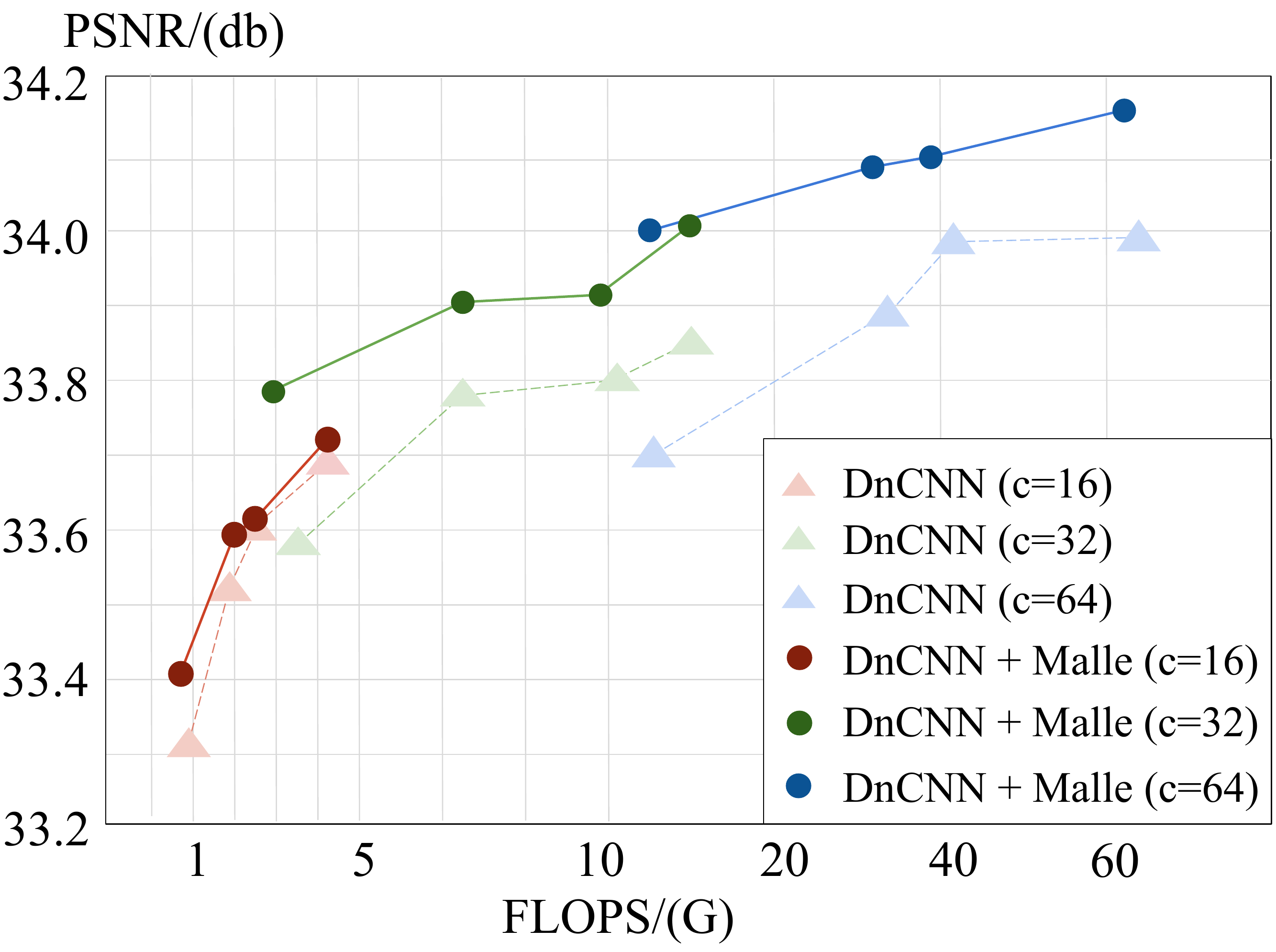} \\
\end{tabular}
}
  \caption{\textbf{PSNR-To-Complexity trade-off of DnCNN and DnCNN with a single \ours}. We build DnCNN-families by setting depth = \{3, 6, 9, 15\} and channel = \{16, 32, 64\}. The three figures from left to right show experiments with $\sigma$ = \{50, 25, 15\}.}
  \label{fig:plug_in_dncnn}
\end{minipage}%
\hfill 
\begin{minipage}[t]{\textwidth}
  \resizebox{0.99\linewidth}{!}{
\begin{tabular}[!t]{ccc}
  \includegraphics[trim={0 0 0 0},clip,width=0.33\linewidth]{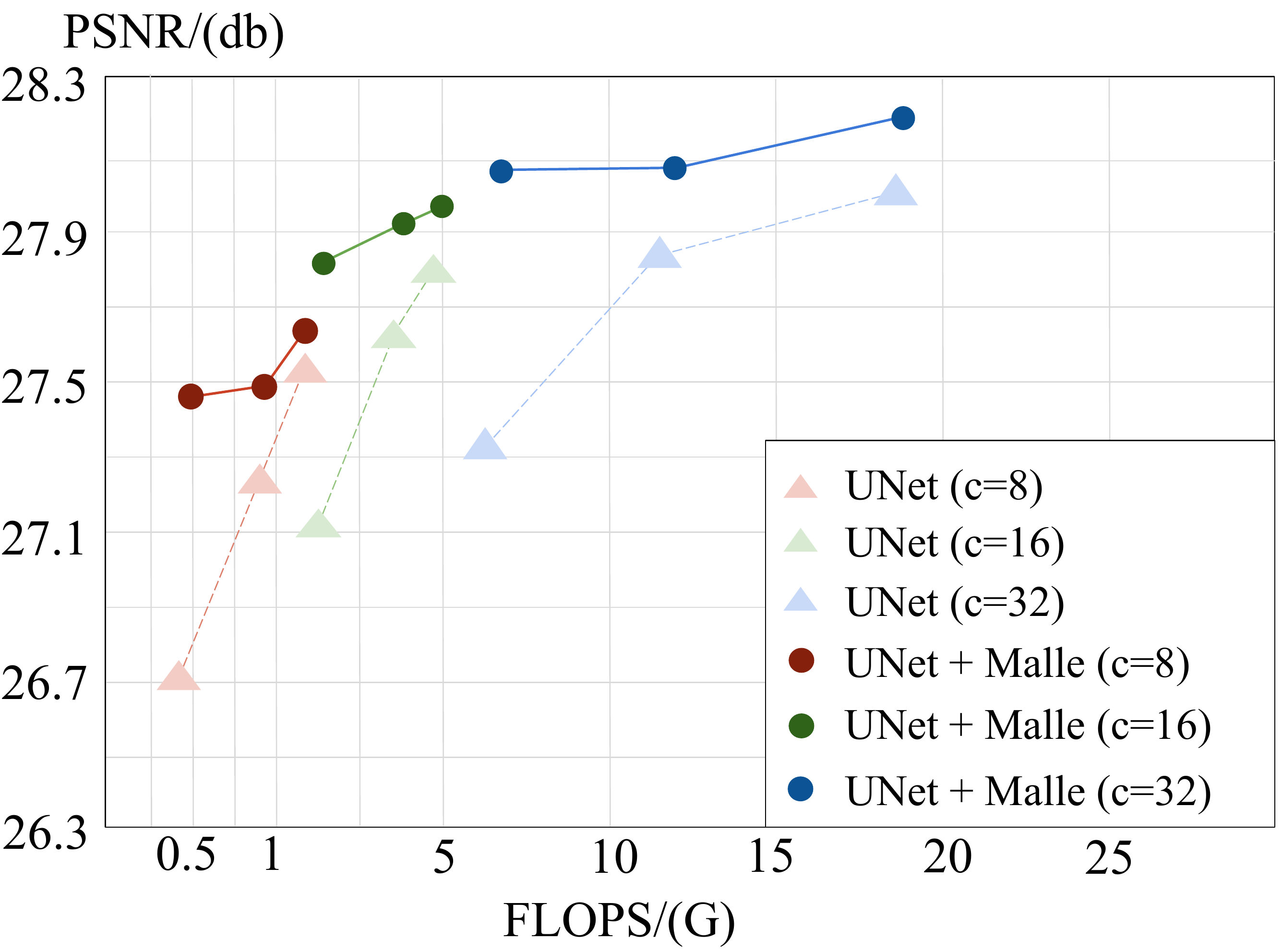} &  
  \includegraphics[trim={0 0 0 0},clip,width=0.33\linewidth]{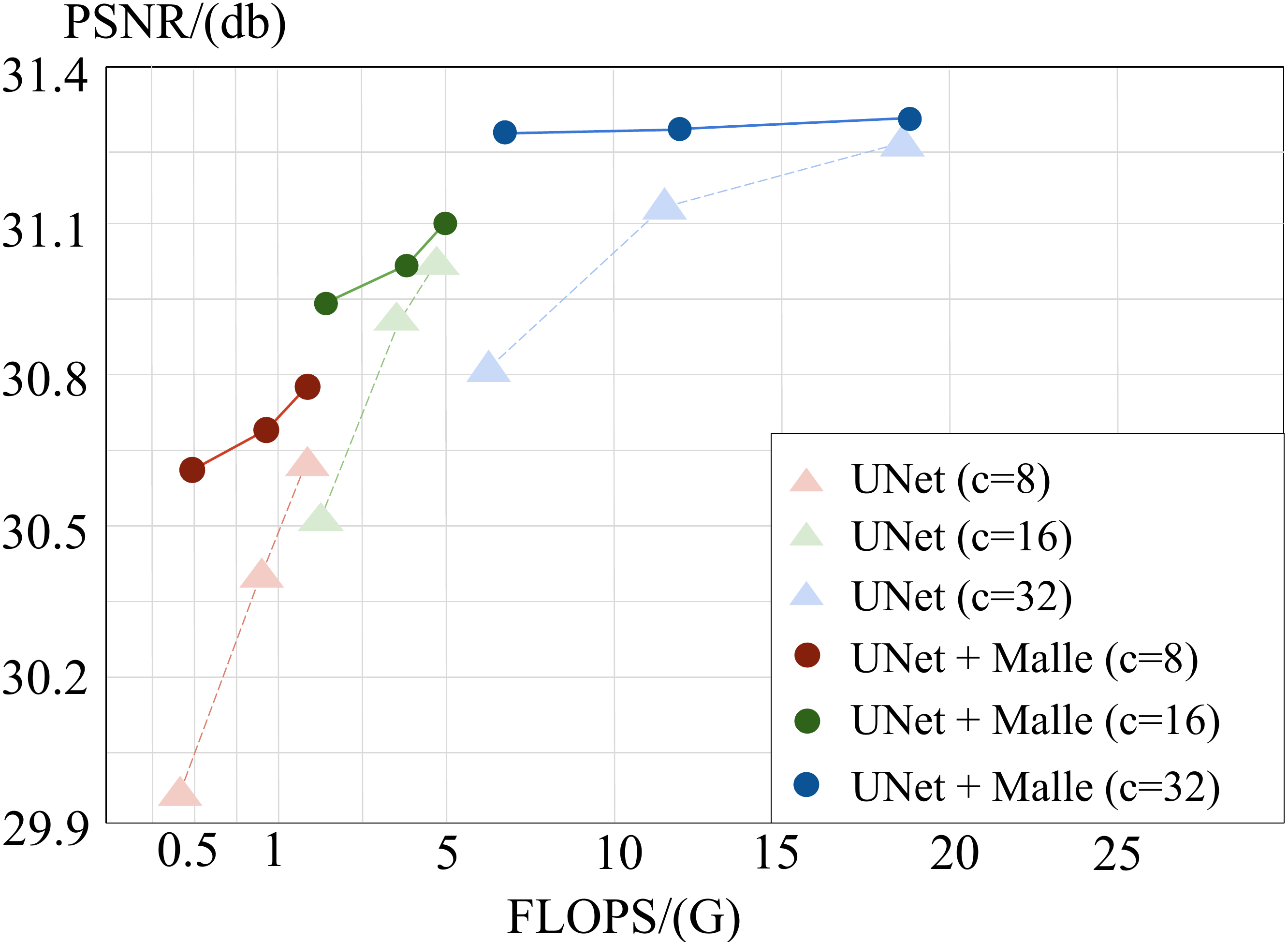} &  
  \includegraphics[trim={0 0 0 0},clip,width=0.33\linewidth]{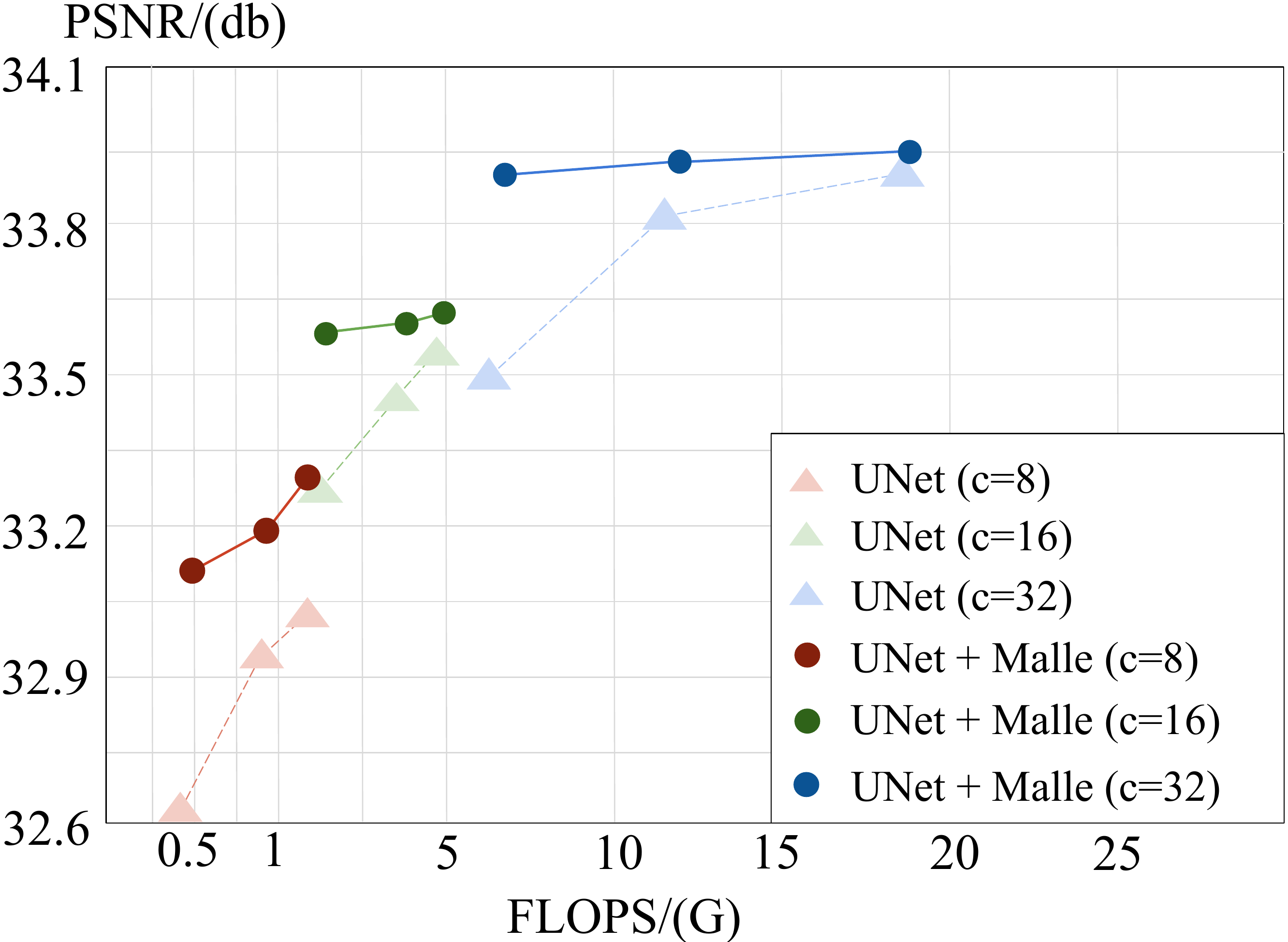} \\
\end{tabular}
}
  \caption{\textbf{PSNR-To-Complexity trade-off of UNet and UNet with a single \ours}. We build UNet-families by setting the encoder-decoder block number = \{2, 3, 4\} and channel = \{8, 16, 32\}. The three figures show experiments with $\sigma$ = \{50, 25, 15\}.}
  \label{fig:plug_in_unet}
\end{minipage}%
\hfill
\begin{minipage}[t]{\textwidth}
  \resizebox{0.99\linewidth}{!}{
\begin{tabular}[!t]{ccc}
  \includegraphics[trim={0 0 0 0},clip,width=0.33\linewidth]{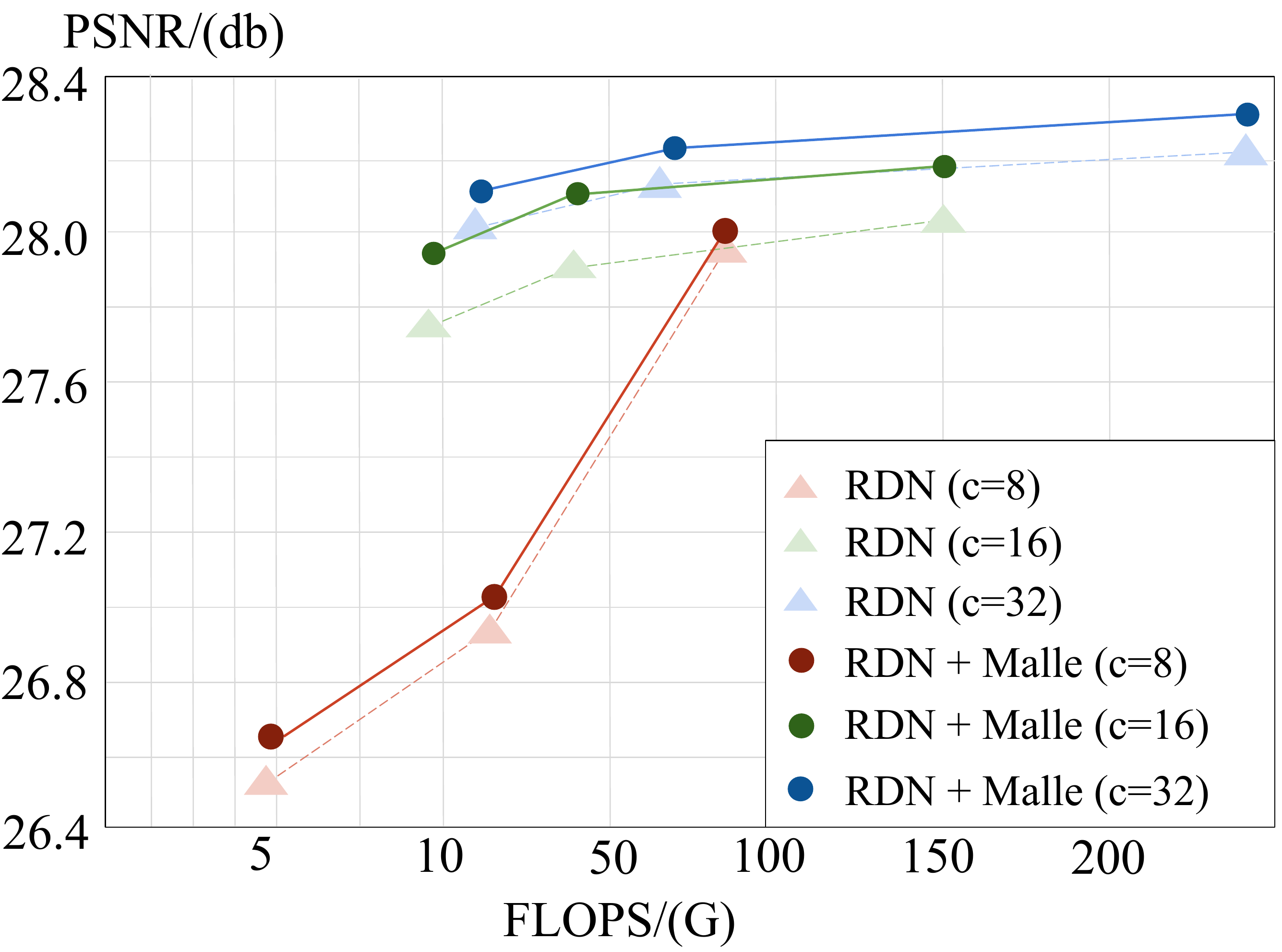} &  
  \includegraphics[trim={0 0 0 0},clip,width=0.33\linewidth]{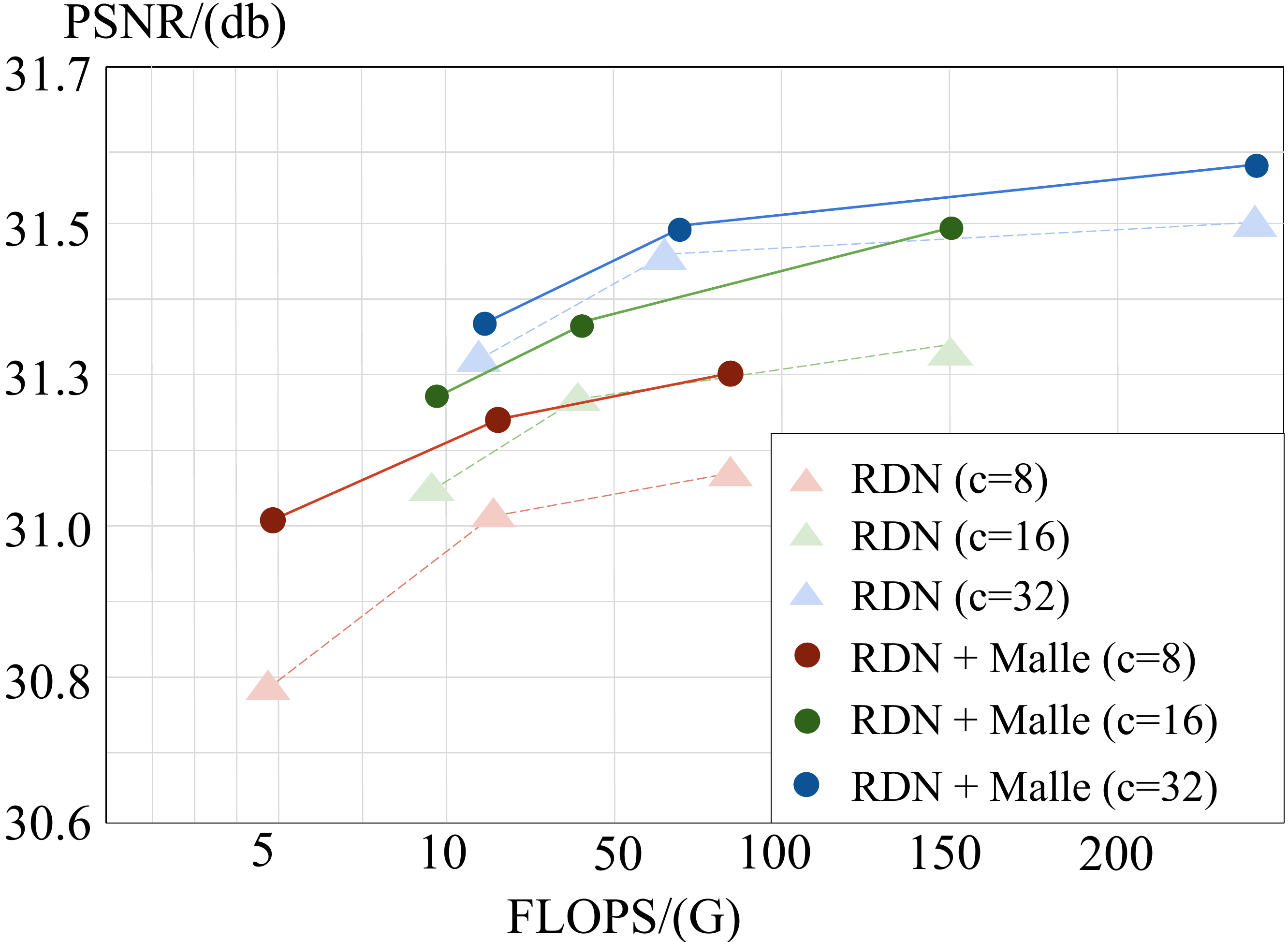} &  
  \includegraphics[trim={0 0 0 0},clip,width=0.33\linewidth]{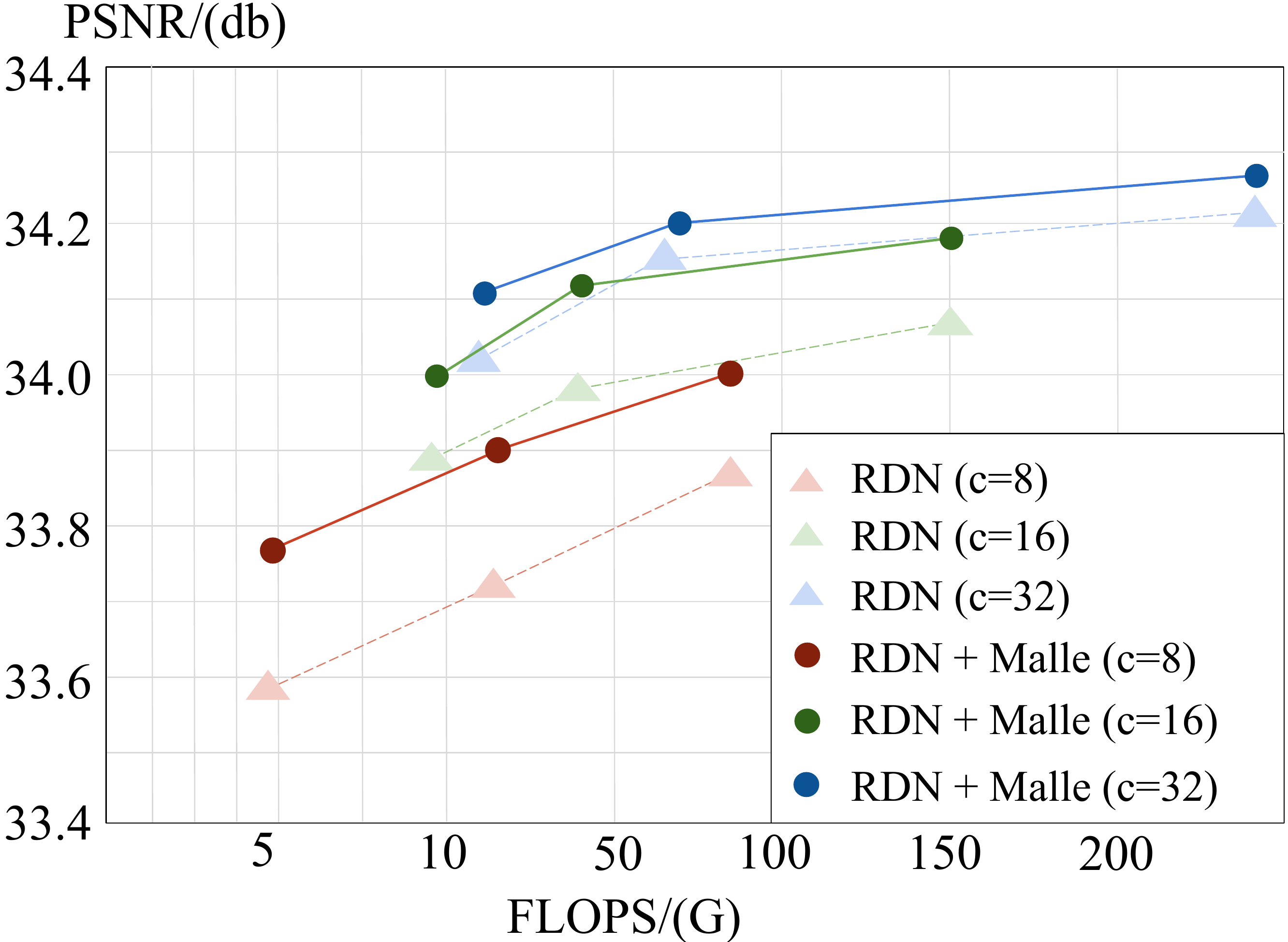} \\
\end{tabular}
}
  \caption{\textbf{PSNR-To-Complexity trade-off of RDN and RDN with a single \ours}. We build RDN-families by setting the residual dense block number = \{3, 6, 10\} and channel = \{8, 16, 32\}. The three figures show experiments with $\sigma$ = \{50, 25, 15\}.}
  \label{fig:plug_in_rdn}
\end{minipage}%
\end{figure*}

\subsection{Comparing MalleConv with Other Dyanmic Kernels}
To demonstrate the efficiency and effectiveness of the proposed MalleConv, we compare specific computational cost and performance of each individual network equipped with MalleConv and other dynamic filters, e.g., HyperNetwork~\cite{ha2016hypernetworks} and Involution~\cite{li2021involution}. We adopt DnCNN~\cite{zhang2017beyond} as our main backbone and replace the middle layer of DnCNN with a single dynamic filter operator. We evaluate three different DnCNN backbones with channel=\{16, 32, 64\}. In each one, the number (depth) of DnCNN backbone are growing from 3, 6, 9, to 15. As shown in Fig.~\ref{fig:hyper_inv_malle}, MalleConv achieves the best performance-efficiency trade-off by significantly improving the PSNR with minimum additional runtime latency. 

\subsection{Comparing with State-of-the-Art Methods}
To fairly compare the runtime speed between MalleNet and other baselines, we train 4 versions of MalleNet: -S, -M, -L, and -XL by increasing the number of channels from 16, 32, 64 to 144. 
We divide evaluated approaches into four categories according to their performance and runtime speed. 
As shown in Table \ref{table:comparison}, on these four categories, MalleNet achieves the best efficiency-performance trade-off and reaches state-of-the-art results among two of our main benchmark test sets. We show the PSNR-to-Complexity trade-off of each method in Fig.~\ref{fig:efficiency_curve} left.

\begin{table}[t]
\small
\begin{center}
\small
\resizebox{0.9\linewidth}{!}{
\setlength{\tabcolsep}{6mm}{
\begin{tabular}{l|c|c|c|c|c} 
\toprule
\multirow{2}{*}{Depth} & \multirow{2}{*}{Metrics}   & \multicolumn{4}{c}{AvgPooling Size}  \\
\cline{3-6}
& & 0 & 2 &  4 & 8 \\

\hline
\multirow{3}{*}{ D=3} & Latency/(ms) & 13.19 & 7.08 & \cellcolor{grey}5.62  &  5.18 \\
& FLOPs/(G) &  43.96 & 18.17 & \cellcolor{grey} 11.71  &  10.10 \\
& PSNR/(dB)   &   27.91    &   \textbf{28.15}    &  \cellcolor{grey}28.07 &  28.01  \\
\hline
\multirow{3}{*}{ D=6}& Latency/(ms) & 17.39 & 11.28 & \cellcolor{grey}9.85  &  9.43 \\
& FLOPs/(G) &  62.05 & 36.26 & \cellcolor{grey}29.81  &  28.19 \\
& PSNR/(dB)   &   28.19    &   \textbf{28.24}    &  \cellcolor{grey}\textbf{28.24} &  28.18  \\
\hline
\multirow{3}{*}{ D=15}& Latency/(ms) &  30.09 & 23.98 &  \cellcolor{grey}22.53 & 22.09  \\
& FLOPs/(G) &  98.24 & 72.44 & \cellcolor{grey} 66.00  &  64.39 \\
& PSNR/(dB)   &   28.25    &   28.28    &  \cellcolor{grey}\textbf{28.31} &  28.28  \\

\bottomrule

\end{tabular}}}
\end{center}
\caption{Ablation study on the size of AvgPooling layer in MalleConv Operator. PSNR results are reported on the CBSD68 test set with $\sigma$ = 50.}
\label{table:downsample}
\end{table}

\subsection{MalleConv Layer with Alternative Backbones} 
To further demonstrate that the proposed Malleable Convolution can benefit wide variety of network architectures, we perform ablation studies by inserting MalleConv into existing well-known backbones as a plug-in operator. Here we choose three popular backbones as our main testbeds. Since most of original network structures are too heavy for edge devices, we also manually build a few cheaper variants by controlling the depth and channel variables. Using DnCNN as an example, the vanilla DnCNN architecture contains 15 layers with 64 channels. We construct its faster version by setting the depth = \{3, 6, 9, 15\} and channel = \{16, 32, 64\}, respectively, and obtain the architecture series of DnCNN with $3 \times 4 = 12$ variants. 

Afterwards, we construct a number of better performing variants of these architecture series, \textbf{by replacing one standard convolution with a single $1 \times 1$ MalleConv operator}. We replace the middle layer of the network with a MalleConv block (detailed architectures are shown in the supplementary material). We conduct experiments on CBSD69 dataset and train these architectures using the same training recipes. As shown in Fig.~\ref{fig:plug_in_dncnn}, \ref{fig:plug_in_unet}, and~\ref{fig:plug_in_rdn}, a single MalleConv block brings significant improvement to all three backbones. 

\begin{table*}[t]
\small
\begin{center}
\small
\resizebox{\linewidth}{!}{
\setlength{\tabcolsep}{4mm}{
\begin{tabular}{l|c|c|c|c|c|c} 
\toprule
\multirow{2}{*}{Method} & \multirow{2}{*}{Latency} & \multirow{2}{*}{FLOPs/(G)} & \multicolumn{2}{c}{SIDD} & \multicolumn{2}{c}{DND} \\
\cline{4-5}  \cline{6-7}
& & & PSNR & SSIM & PSNR & SSIM \\
\hline
DnCNN~\cite{zhang2017beyond} & 21.69 & 68.15 & 23.66 & 0.583 & 32.43 & 0.79 \\
BM3D~\cite{dabov2007image} & 41.56 & -  & 25.78 & 0.685 & 34.51 & 0.851 \\
WNNM~\cite{gu2014weighted} & - & - & 25.78 & 0.809 & 34.67 & 0.865 \\
CBDNet~\cite{guo2019toward} & - & - & 30.78 & 0.754 & 38.06 & 0.942 \\
RIDNet~\cite{anwar2019real} & 98.13 & - & 38.71 & 0.914 & 39.26 & 0.953 \\
VDN~\cite{yue2019variational} & - & - & 39.28 &  0.909 & 39.38 & 0.952\\
ACDA~\cite{wang2021adaptive} & -& - & 39.32 & 0.912 & - & -\\
MPRNet~\cite{zamir2021multi} & - & 573.50 &  39.71 & 0.958 & 39.80 & 0.954 \\
NBNet~\cite{cheng2021nbnet} & 37.44 & 88.70 & 39.75 & 0.973 & 39.89 & 0.955\\
MIRNet~\cite{zamir2020learning} & 192.61 & 787.04 & 39.72 & 0.959 & 39.88 & 0.956\\
HINet~\cite{chen2021hinet} & 32.83 & 170.71 & 39.99 & 0.958 & - & - \\
\textbf{MalleNet-R} & \textbf{13.58} & \textbf{29.11} & 39.56 & 0.941 & 39.21 & 0.949 \\
\bottomrule
\end{tabular}}}
\end{center}
\caption{\textbf{Comparing MalleNet with the State-of-the-art methods on real-world benchmark SIDD and DND.}  We try our best to use the official implementation provided by the authors to calculate the FLOPs cost and runtime speed.}
\label{table:sidd}
\end{table*}


\subsection{Visual Comparison and Interpretation}

We first compare our best architecture MalleNet-XL with previous state-of-the-art approaches~\cite{zhang2017beyond,zhang2018residual,liang2021swinir,liang2021swinir}, as shown in Fig. \ref{fig:visual_comparison}. The examples produced by MalleNet preserve rich details and impressive textures while saving up to $\times$\textbf{8.91} inference time compared to the best baseline, further demonstrating the effectiveness of our approach. Moreover, in the ``ultra-fast'' setting, we decrease the depth of DnCNN from 15 to 3 to obtain a much faster variant of DnCNN architecture. However, the image quality also degrades as shown in the bottom-left of Fig.~\ref{fig:dncnn_visual_comparison}. 
In contrast, when replacing the middle layer of DnCNN with a single $1 \times 1$ MalleConv operator (DnCNN w/ MalleConv), it uses slightly more
computational time, but achieves significantly better visual quality, as shown in the bottom-right of Fig.~\ref{fig:dncnn_visual_comparison}.

\begin{figure*}[t]
\centering
\includegraphics[width=1.0\linewidth]{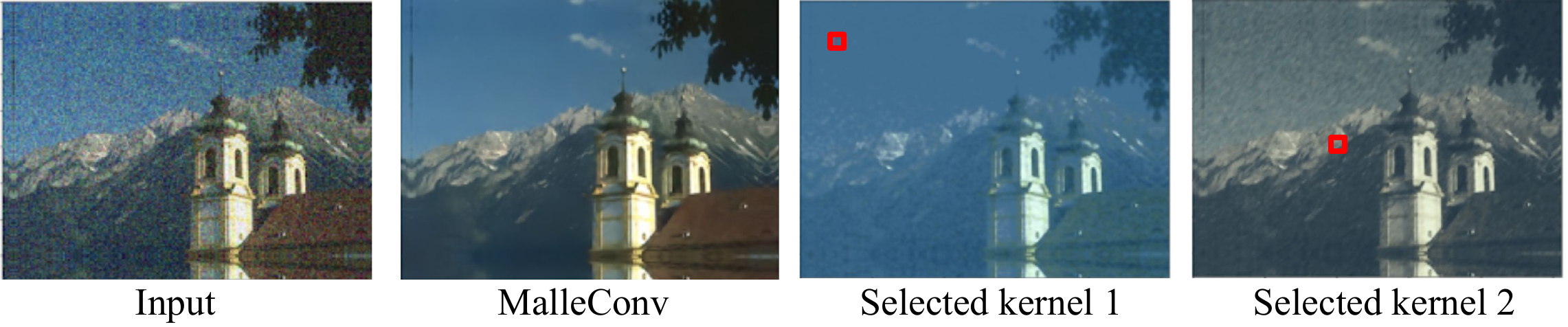}
\caption{Comparison between default MalleConv output (column2) and outputs using two selected kernels (column 3 and 4).}
\label{fig:grid_visualization}
\end{figure*}

Furthermore, to illustrate how spatially-varying kernels in MalleConv capture heterogeneous visual patterns, we replace the spatially varying kernels in MalleNet with one selected kernel and apply it to the entire image. Fig.~\ref{fig:grid_visualization} compares the default output of MalleNet (column 2) with the one that applies a selected kernel (columns 3 and 4). When a kernel generated from a sky region (column 3) is applied, the network is observed to denoise the rest of the image as if they are the sky. Similarly, using a kernel from a snowy-mountain patch will generate output that looks like snowy mountain (column 4). By combining kernels that are dedicated to different local image statistics together, MalleConv can better model the heterogeneous spatial patterns and yield better results.

\label{sec:memory_analysis}
\subsection{Analysis of Runtime Latency and Memory Cost}
\begin{wrapfigure}{r}{60mm}
\centering
\includegraphics[width=0.5\textwidth]{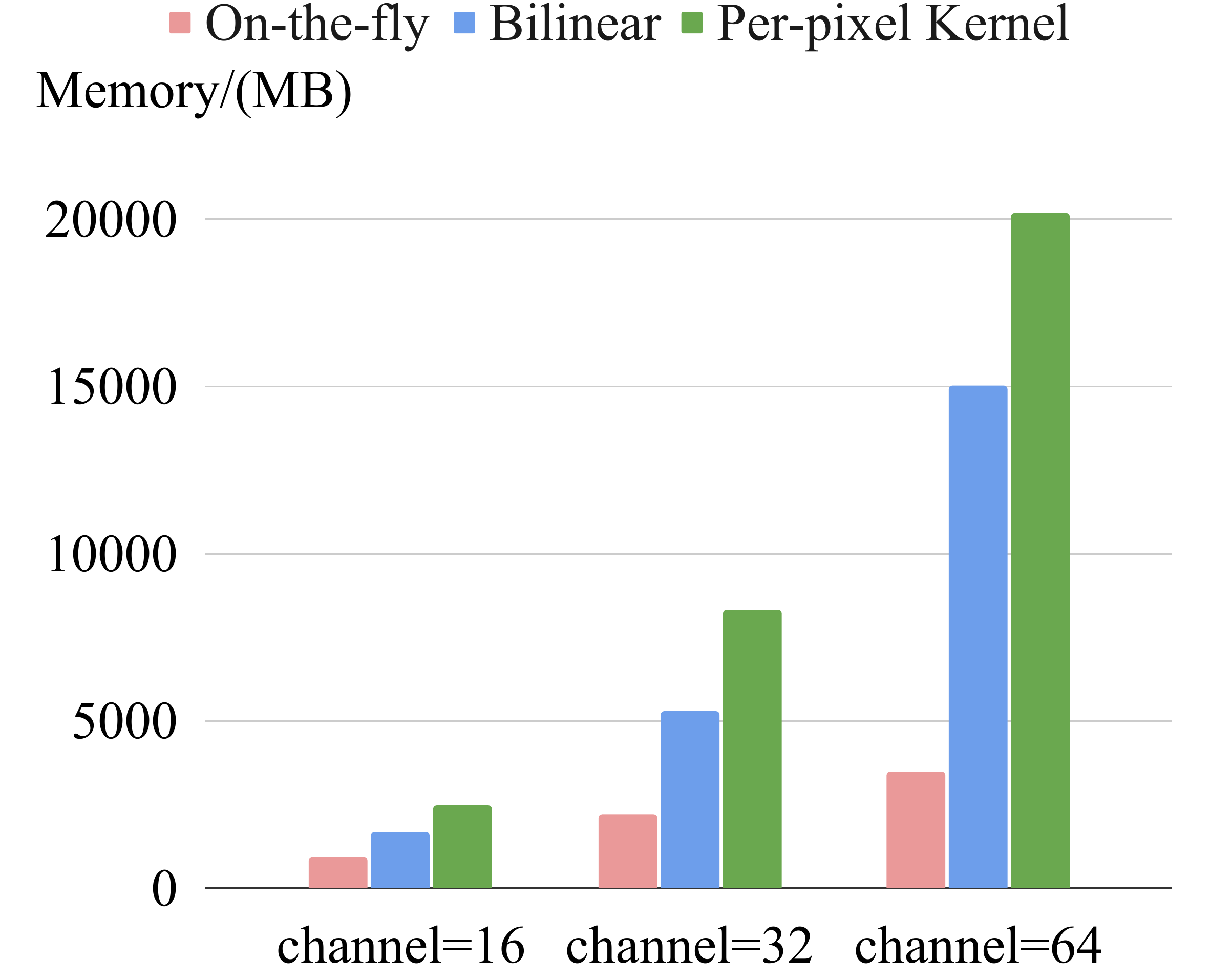}
\par
\caption{Memory cost comparison between the proposed method and per-pixel kernel prediction approaches (HyperNetwork).\label{fig:memory}}
\vspace{-0.5em}
\end{wrapfigure}

In Fig.~\ref{fig:memory}, we compare the memory cost of each operator during the training process. We conduct our testbed on three different modules: 1) The $1 \times 1$ MalleConv with input and output channel to be {16, 32, 64}. 
MalleConv generates smaller-size of dynamic kernels and then applies it back to the full-resolution feature using on-the-fly slicing operator; 
2) We directly upsample the generated dynamic kernel via an $8\times$ bilinear upsampling operator, to match the resolution of input features; 3) We remove the downsampling and maxpooling layers in the proposed efficient prediction network, thus it will generate per-pixel dynamic kernels and apply it to the feature map. As shown in Fig.~\ref{fig:memory}, the memory cost of MalleConv is much smaller than other two counterparts, since it only needs to predict a smaller-size of filters compared to the per-pixel kernel prediction methods, and does not store the intermediate feature map of upsampled kernels compared to the bilinear interpolation operator.

Moreover, we conduct the ablation study on the downsampling ratio of the proposed efficient predictor network. Similar to the aforementioned setting, we set our testbed on DnCNN approach and examine three different architectures by setting depth = \{3, 6, 15\}. We evaluate the runtime speed, FLOPs cost, and PSNR value of four variants with the size of the AvgPooling layer equal to \{0, 2, 4, 8\}. As shown in Table~\ref{table:downsample}, by processing a $4\times$ downsampled feature map, our proposed efficient predictor network achieves a ``win-win'' in terms of both performance and efficiency. This demonstrates that applying the prediction network on a lower resolution feature map can not only improve the performance, due to a larger receptive field, but also save computations



\subsection{Evaluation on Real Sensor Noise}
To further demonstrate the generalization ability of MalleNet, we evaluate our approaches to real sensor noise. Similar to previous works~\cite{cheng2021nbnet,zamir2020learning}, we adopt Smartphone Image Denoising Dataset (SIDD)~\cite{abdelhamed2018high} and Darmstadt Noise Dataset (DND)~\cite{plotz2017benchmarking} as main benchmarks.
We use training data from SIDD as our training set and evaluate our method on both two test sets. In the training process, We adopt Adam Optimizer with a batch size of 128, the weight decay is set to 0.03, and the learning rate is set to 2e-4. We randomly crop $256 \times 256$ patches and apply random rotation and flipping. We train a real denoiser MalleNet-R, by slightly modifing the channel/depth of MalleNet-M architecture and replace Inverse Bottleneck Block with standard residual block (see supplementary materials for details). As shown in Table ~\ref{table:sidd}, MalleNet-R achieves lower latency (\textbf{13.58 ms}) compared with other methods. In terms of image quality, MalleNet-R is able to reach similar PSNR/SSIM compared to most baselines, and only slightly behinds the approaches with very heavy computational cost or equipped with complex channel/spatial attention module. We show the PSNR-to-Complexity trade-off of each method in Fig.~\ref{fig:efficiency_curve} right. More visual comparisons are included in the supplementary materials.

\begin{figure*}[t]
\center
\begin{minipage}[t]{\textwidth}
  \resizebox{0.99\linewidth}{!}{
\begin{tabular}[!t]{cc}
\includegraphics[width=0.4\linewidth]{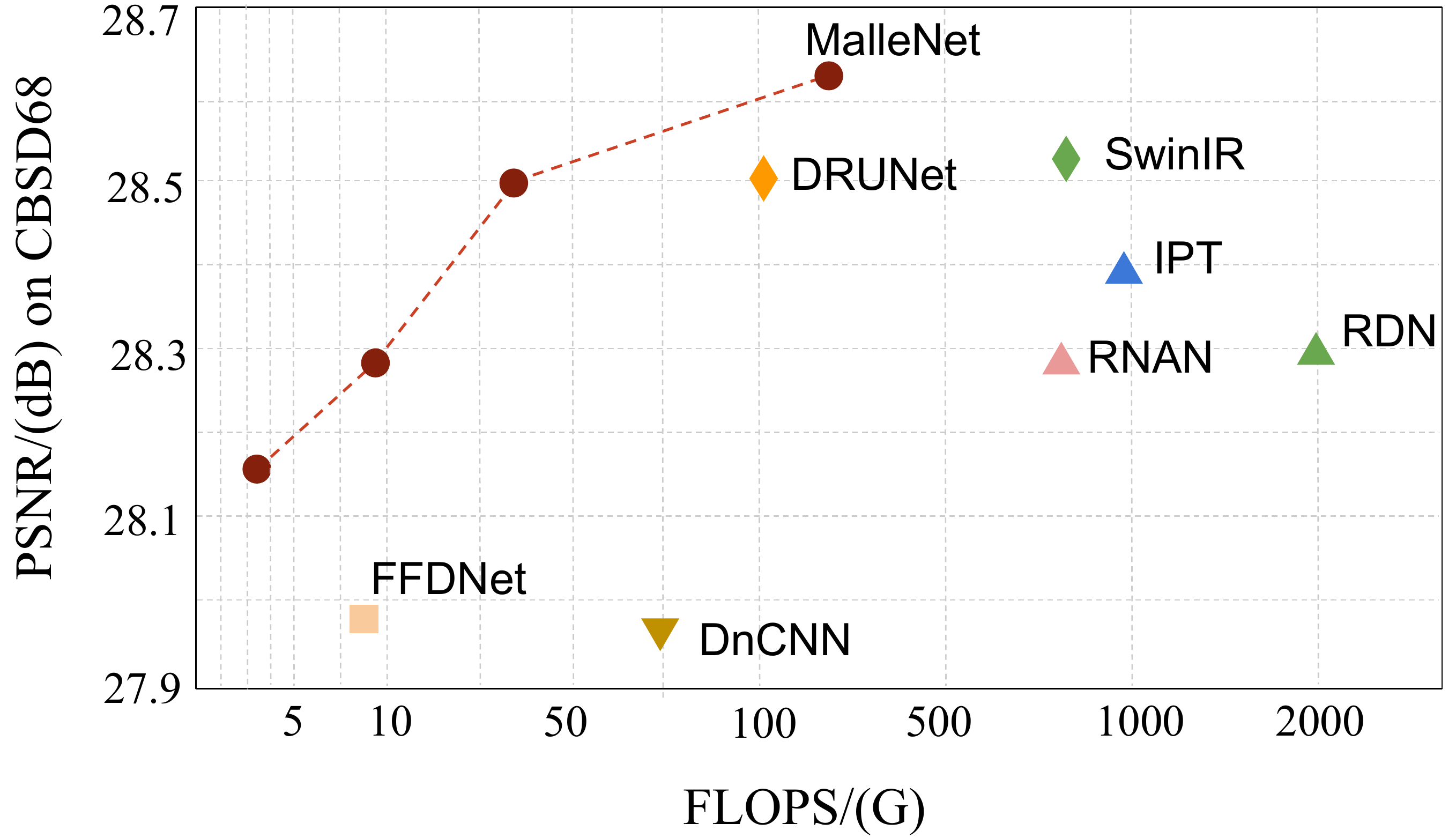} & \includegraphics[width=0.4\linewidth]{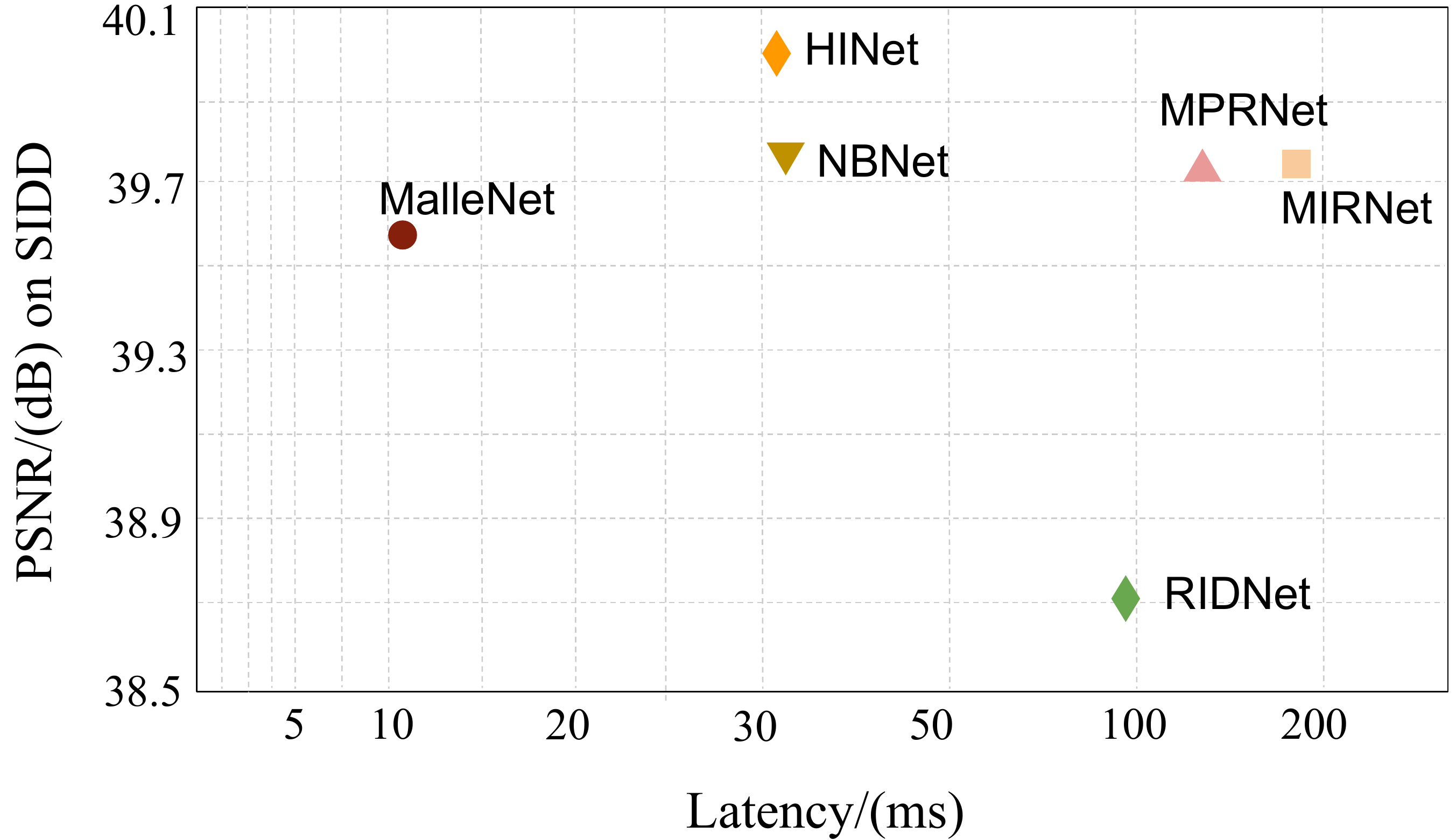}
\end{tabular}
}
\caption{\textbf{Results on CBSD68 test set ($\sigma$ = 50) and SIDD validation set.} Our proposed MalleNet architecture achieves a better trade-off between quality and speed.}
  
\label{fig:efficiency_curve}
\end{minipage}%
\end{figure*}

\section{Conclusions}
In this work, we propose Malleable Convolution (MalleConv), an efficient variant of spatially-varying convolution tailored for ultra-fast image denoising. MalleConv processes a low-resolution feature map and generates a much smaller set of spatially varying filters. The generated filters inherently fit the heterogeneous and spatially varying patterns presented in natural images, while taking little additional computational costs.
Despite its effectiveness, we also observe that very deep or wide architectures benefit less from MalleConv, as they may also capture heterogenous image statistics in a less efficient way. Although in this work, we only evaluated MalleConv on image denoising, we believe MalleConv is also capable in other image processing tasks, like dehazing. Another future work is to combine MalleConv with attention mechanism~\cite{liang2021swinir} or deformable shape~\cite{dai2017deformable} to further improve its quality in applications with less computational constraints.

\section{Acknowledgement}
We would like to express our gratitude to the Google Research Luma team, in particular Zhengzhong Tu for generously providing us with the concrete training recipes on real-world denoising benchmarks.
\newpage
{\small
\bibliographystyle{ieee_fullname}
\bibliography{egbib}
}

\end{document}